\documentclass[11pt]{article}
\usepackage{amsmath}
\usepackage{algorithm,algorithmic}
\usepackage{times}
\usepackage{graphicx}
\usepackage{multirow}

\usepackage{bm}
\usepackage{amssymb}

\usepackage{rotating}
\usepackage{cite}
\usepackage{bbm}
\usepackage{latexsym}
\usepackage{fancyhdr}

\begin{document}
\hspace{13.9cm}1

\ \vspace{20mm}\\

{\begin{center}
{\LARGE Sparse Coding Approach for Multi-Frame Image Super Resolution}

\ \\
{\bf \large Toshiyuki Kato$^{1}$, Hideitsu Hino$^{2}$, and Noboru
  Murata$^{1}$} \\
$^{1}$Waseda University, 3-4-1 Ohkubo, Shinjuku,
Tokyo, Japan,\\
$^{2}$University of Tsukuba.\\
1-1-1 Tennodai, Tsukuba, Ibaraki, 305--8573, Japan
\end{center}}

{\bf Keywords: 
image super-resolution, multi-frame super-resolution, sparse coding}
\thispagestyle{fancy}
\rhead{}
\lhead{}

%Abstract
\begin{center} {\bf Abstract} \end{center}
An image super-resolution method from multiple observation of
 low-resolution images is proposed. The method is based on sub-pixel
 accuracy block matching for estimating relative displacements of
 observed images, and sparse signal representation for estimating the
 corresponding high-resolution image. 
 Relative displacements of small patches of observed low-resolution
 images are accurately estimated by a computationally efficient block
 matching method. Since the estimated displacements are also regarded as a
 warping component of image degradation process, the matching
 results are directly utilized to generate low-resolution dictionary for
 sparse image representation. The matching scores of the block matching
 are used to select a subset of low-resolution patches for
  reconstructing a high-resolution patch, that is, 
 an adaptive selection of informative
 low-resolution images is realized. 
 When there is only one low-resolution image, the proposed method 
 works as a single-frame super-resolution method.
 The proposed method is shown to perform comparable or superior to 
 conventional single- and multi-frame super-resolution methods through experiments using
 various real-world datasets.
%%%%%%%%%%%

\section{Introduction}
Super-resolution (SR)  have been receiving a large amount of
attention for creating clear images from low-resolution images.
In stark contrast to simple picture interpolation techniques, SR methods
utilize prior knowledges or assumptions on the structure and
relationship of high- and low-resolution images. 
Development and spread of video equipments drive
 a growing need for SR techniques, which enable us to create
high-resolution (HR) images from low-resolution (LR) images. 
See, e.g., \cite{Borman98,Tian2011} for comprehensive surveys.

Super-resolution techniques can be divided into two categories,
reconstruction-based SR~\cite{mapsr,frsr,kanemura}, and example-based
SR~\cite{freeman1,chang1,kamimura,sun}. 
In the former approach, we compute HR images by simulating the image formation
process.
It is often used for multi-frame SR, where multiple LR images are used to obtain an HR image.
A random Markov field model is usually adopted to
represent the relationship between pixels of LR and HR images. This
approach is intuitive and natural for SR. It is shown to give favorable
results with an appropriate prior~\cite{Katsuki2012}, but it often requires
vast amounts of computation.

Example-based SR aims at inferring HR images based on small image
segments extracted from training HR images.
In many cases, it is adopted when we can use only one LR
 image. One of the representative works of example-based SR is the method by
 Freeman et al.~\cite{freeman1}, which is based on Neighborhood Embedding{\cite{LLE2000}}.
Recently, Yang et al~\cite{yang1,yang2} improved~\cite{freeman1} and proposed an SR method based on sparse
 coding~{\cite{olshausen1}}.
Sparse coding is a methodology to represent observed signals with combinations
of only a small number of basis vectors chosen from a large number of
 candidates. A set of basis vectors is called {\it{dictionary}}. 
A lot of single frame SR methods based on sparse coding are proposed, and they are
 experimentally shown to offer favorable HR images.
In SR based on sparse coding, a dictionary for HR images is prepared in
 advance, and taking into account the image degradation process, a
 dictionary for LR images is generated from the HR dictionary.
An LR image is represented by a combination of LR bases, then the
 coefficients for LR bases are used for combination coefficients of HR
 bases to obtain the sharpened image.
In~\cite{yang1,yang2}, sparse coding is ingeniously utilized to obtain
 an HR image from only one LR observation. However, it should be possible
 to gain further improvements when we have multiple LR images for
 reconstructing an HR image.

In this paper, we propose a multi-frame SR method based on sparse
coding.
In general, multi-frame SR methods require registration of LR images,
that is, we have to estimate relative displacements of LR images.
We propose to use a sub-pixel accuracy block matching method for image
registration.
 One of the contributions of the proposed method
is in
using the results of block matching to generate LR dictionaries.
 Another contribution is that the proposed
method can adaptively select informative LR images for constructing HR
image, which is also a beneficial side-effect of sub-pixel accuracy
block matching. 
In that sense, the proposed method is {\it{adaptively}} selects
the patches to be used for SR. 
By thresholding the matching score, the number of LR
images used for reconstructing HR image varies in each small
patch. We note that when there is only one LR image, the proposed method
is the same as the conventional SR method based on sparse coding.

The rest of this paper is organized as follows. 
The image observation model is introduced in section~II, and the
sub-pixel accuracy block matching method is briefly explained without
technical details in section~III.
The notion of sparse coding is introduced in section~IV.
 In section~V, conventional super-resolution methods based on sparse
coding is explained, and in section~VI, a novel multi-frame SR method based on sparse
coding is proposed. 
Section~VII shows experimental results, and the last section
is devoted to concluding remarks.

\section{Image Observation Model}
\label{sec_model}
In this section, we describe the image observation model. 
Following the idea of~\cite{frsr}, we assume a continuous image $\tilde{X}(x,y)$ where $(x, y) \in \mathbb{R}^{2}$ are coordinate values. Then,
 we assume that an ideal discrete HR image $\mathbf{X}$ and an LR image $\mathbf{Y}$ are 
 sampled from the continuous image $\tilde{X}$ according to the following model:
\begin{align}
\label{single_X}
X[m,n]&= \left[\mathcal{W} \left( \Tilde{X}(x,y) \right) \right] \downarrow_X \\
\label{single_Y}
Y[m,n] &= \left[\mathcal{H} * \mathcal{W} \left( \Tilde{X}(x,y) \right) \right] \downarrow_Y + \mathcal{E}[m,n],
\end{align}
where $\mathcal{W}$ and $\mathcal{H}$ are warp and blur operators,
$\downarrow_{X}$ and $\downarrow_{Y}$ are quantization operators to
generate HR and LR images, and $\mathcal{E}$ is an additive noise. In
this paper, we denote coordinates in continuous space and discrete space
by $(x,y)$ and $[m,n]$, respectively. The blurring is expressed by the
convolution operator $\ast$. 
 
Following the conventional formulation of super resolution, we treat the HR and LR images as vectors $X \in \mathbb{R}^{p_{h}}, Y \in \mathbb{R}^{p_l}$, 
and the LR observation is assumed to be related with the HR image by
\begin{equation}
\label{single_dgp}
\mathbf{Y} = SHW\mathbf{X} + \bm{\varepsilon},
\end{equation}
where $\mathbf{Y}$ and $\mathbf{X}$ are column vectors with stacked images, 
 the matrix $W \in \mathbb{R}^{p_{h} \times p_{h}}$ encodes the warping or spacial distortion,
  the matrix $H \in \mathbb{R}^{p_{h} \times p_{h}}$ models the blurring effect,
$S \in \mathbb{R}^{p_{l} \times p_{h}}$ is the down-sampling
operator, and $\bm{\varepsilon} \in \mathbb{R}^{p_{l}}$ is the Gaussian noise term. 

Example-based SR approaches usually extract small patches
$\mathbf{x} \in \mathbb{R}^{q_{h}}$ from the HR image and
$\mathbf{y} \in \mathbb{R}^{q_{l}}$ from the LR
image $\mathbf{Y}$. 
The whole HR image $\mathbf{X}$ is obtained by integrating HR patches
$\mathbf{x}$. 
With some abuse of notation of the operators and the noise term, each patch pair
$(\mathbf{x}, \mathbf{y})$ is connected by the observation model
\begin{align}
\label{eq:Smodel_patch}
\mathbf{y} = SHW\mathbf{x} + \bm{\varepsilon},
\end{align}
where $W \in \mathbb{R}^{q_h \times q_h}$, $H \in \mathbb{R}^{q_h \times
q_h}$, $S \in \mathbb{R}^{q_l \times q_h}$, and $\bm{\varepsilon}
\in \mathbb{R}^{q_l}$. 
The blurring effects are often modeled by convolution with a point
spread function. In this paper, we use $9 \times 9$ Gaussian filter for
the point spread function.
The down-sampling process $S$ is assumed to be an impulse sampling.
The down-sampled image is further affected by the sensor noise and color
filtering noise, and they are represented by an additive Gaussian noise term. 
In general, the spacial distortion $W$ includes translation, rotation,
deformation and other possible distortions. In this paper, we restrict 
$W$ to simple translations, which is reasonable when we treat small
patches instead of the whole image. 
In multi-frame SR, we assume that a set of
 $N$ observed LR
images $\mathbf{Y}_1,\mathbf{Y}_2,\cdots,\mathbf{Y}_N$ are given. 
We first choose a {\it{target}} LR image
$\mathbf{Y}_{t}$ out of $N$ observed images, and the final
output of an SR method is the HR version of $\mathbf{Y}_{t}$. We refer
other LR images as {\it{auxiliary}} LR images henceforth. Without loss
of generality, we let $t=1$, i.e., ${\mathbf{Y}}_{1}$ is the
target image and ${\mathbf{Y}}_{j}, j=2,\dots, N$ are the auxiliary images.
The LR observations are related with the HR image
$\mathbf{X}$ by
\begin{align}
\mathbf{Y}_{j} = S_{j}H_{j}W_{j}\mathbf{X} + \bm{\varepsilon}_{j}, \quad
 j=1,\dots,N,
\label{eq:SRmodel}
\end{align}
where $W_{j} \in \mathbb{R}^{p_{h} \times p_{h}}$ encodes relative
displacement of a auxiliary image ${\mathbf{Y}}_{j}$ to the target
image ${\mathbf{Y}}_{1}$ in the HR image coordinate, and $W_{1}$ is the identity operator. The operator
$H_{j} \in \mathbb{R}^{p_{h} \times p_{h}}$ models the blurring effect,
$S_{j} \in \mathbb{R}^{p_{l} \times p_{h}}$ models the down-sampling,
and $\bm{\varepsilon}_{j} \in \mathbb{R}^{p_{l}}$ is the Gaussian noise
term.

\section{Registration by Block Matching}  % - - - - - - - - - - - - - - - - - - - - - - - - - - - - - -
\begin{figure}[t]
  \begin{center}
  \includegraphics[clip,height=5cm]{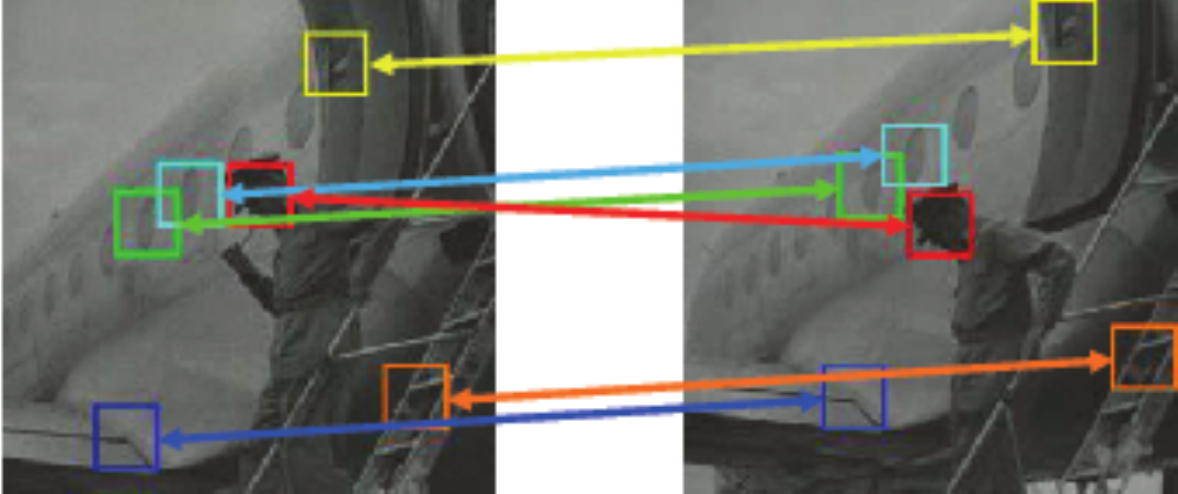}
  \end{center}
 \caption{Conceptual figure of block matching}
\label{bm}
 \end{figure}

\label{sec_bm}
In multi-frame SR, a reliable motion estimation
is essential for achieving high-quality HR image. Image registration means not only
 identifying relative displacements between observed LR images, but also means
the estimation of translation operators $W_{1},\dots,W_{N}$ in the
model~\eqref{eq:SRmodel}.
Various motion estimation methods have been developed. In this paper, we
adopt the block matching method~\cite{BM1,BM2} as shown in Fig.~\ref{bm}, because the sparse coding
method also extracts small patches from images.
As mentioned earlier, we assume that positional relationship of blocks
are fully described by parallel translation. 
Multi-frame SR techniques achieve remarkable results in resolution enhancement by estimating the sub-pixel shifts between images.
There are a variety of strategies for estimating positional
relationship in sub-pixel
accuracy~{\cite{match1,match2,ruiji1,ruiji2,match3,match4}}. In this paper, 
we adopt the 
two-parameter simultaneous estimation method for similarity
interpolation~\cite{2dest2}, 
which balances accuracy and computational efficiency. 
By the block matching method, we obtain the similarity score 
\begin{equation}
\label{eq:simscore}
\text{sim}({\bf{y}}_{t},{\bf{y}}_{j}) =
\frac{ {\bf{y}}_{t}^{\top}{\bf{y}}_{j}}
{\|{\bf{y}}_{t}\|_{2} \cdot
 \|{\bf{y}}_{j} \|_{2} } \in [0,1]
\end{equation}
between patches from the target image ${\bf{Y}}_{1}$ and patches from
auxiliary images ${\bf{Y}}_{j}$. Later, these scores are used to judge whether
the patches extracted from auxiliary images should be used for the HR image
reconstruction or not.

\section{Sparse Coding} % - - - - - - - - - - - - - - - - - - - - - - - - - - - - - -
\label{sec_sc}
Sparse coding~\cite{olshausen1,OLS05,Nat1995,elad2010sparse}
is a methodology to represent observed signals with combinations
of only a small number of basis vectors chosen from a large number of candidates.
These basis vectors will be called {\it atoms} henceforth. 

Let $\mathbf{D} = [\mathbf{d}_1,\mathbf{d}_2,\dots,\mathbf{d}_K] \in
\mathbb{R}^{p \times K}$ be a dictionary which consists of $K$ different
atoms, ${\boldsymbol \alpha \in \mathbb{R}^{K}}$ be the coefficient
vector for sparse representation of the signal $\mathbf{s} \in
\mathbb{R}^p$. Then, the $\ell^{0}$-norm sparse coding problem is formally written as 
\begin{align}
\label{eq:sc1}
\text{minimize} \hspace{5pt} \|{\boldsymbol \alpha}\|_0 \hspace{10pt}
 \text{subject to} \hspace{10pt} \|\mathbf{D}{\boldsymbol \alpha -
 \mathbf{s}}\|_2^2 \leq \epsilon,
\end{align}
where $\epsilon$ is a predefined tolerance for approximation, or 
\begin{align}
\label{eq:sc2}
\text{minimize} \hspace{5pt} \|\mathbf{D}{\boldsymbol \alpha -
 \mathbf{s}}\|_2^2 \hspace{10pt} \text{subject to} \hspace{10pt}
 \|{\boldsymbol \alpha}\|_0 \leq \mathrm{T},
\end{align}
where $T$ is a predefined number of atoms to be used for
representation. The $\ell^{0}$-norm $\| \boldsymbol{\alpha}\|_{0}$ is defined
by the number of nonzero elements of a vector $\boldsymbol \alpha$
\begin{equation}
 \| \boldsymbol{\alpha}\|_{0} =
\# \{ k | \alpha_{k} \neq 0, \, k \in \{1,\dots,K\} \}.
\end{equation}
Given a dictionary $\bf D$, the problems \eqref{eq:sc1} and
\eqref{eq:sc2} are known to be {\it NP-hard}~\cite{mallat97},
and the matching pursuit type algorithms~{\cite{mp1,omp1,omp2}} are
often used for obtaining approximate solutions.
As another approach for sparse coding, the problem is relaxed to an
$\ell^1$-norm minimization problem as
\begin{align}
\text{minimize} \hspace{5pt} \|{\boldsymbol \alpha}\|_1 \hspace{10pt}
 \text{subject to} \hspace{10pt} \|\mathbf{D}{\boldsymbol \alpha -
 \mathbf{s}}\|_2^2 \leq \epsilon.
\label{eq:bp}
\end{align}
This problem~\eqref{eq:bp} adopts the $\ell^1$-norm of coefficients as a
measure of sparsity, and referred to as the $\ell^1$-norm sparse
coding. Using the Lagrange multiplier $\eta$, this problem is rewritten as
\begin{align}
\text{minimize} \hspace{5pt} \frac{1}{2}\| \mathbf{D}\boldsymbol \alpha
 - \mathbf{y}\|^2_2 + \eta \| \boldsymbol \alpha \|_{1}
\label{eq:l1SC}
\end{align}
where $\eta \geq 0$ represents the strength of the sparseness constraint.
 In~\cite{bpdn}, it is argued that the $\ell^{1}$-norm
sparse coding is more stable than the $\ell^{0}$-norm sparse coding.
The problem~\eqref{eq:l1SC} is the same form as Lasso~{\cite{lasso}},
which is widely used sparse linear regression, and there are some
computationally efficient algorithms for estimating the coefficient such
as LARS-lasso algorithm~{\cite{lars}} and feature-sign search algorithm~{\cite{esc}}.
 In this paper, we adopt the $\ell^{1}$-norm sparse coding as a building
block for super-resolution, and use the LARS-lasso algorithm for solving
the problem~\eqref{eq:l1SC}.

In sparse coding, the design of dictionary $\bf D$ is of prime importance. 
The first attempt of the example-based SR method employs small patches of HR images themselves
as atoms~\cite{freeman1}, though, learning the dictionary from observed data is shown to
improve the reconstruction performance. Actually, the HR and LR dictionary
learning is the core
technical component of the state-of-the-art SR methods based on sparse coding~\cite{yang1,yang2,elad1}.
 Representative methods for dictionary learning are Method of
Optimal Directions~\cite{engan99} which is based on vector quantization,
and K-SVD~\cite{ksvd} which is based on k-means clustering and the
singular value
decomposition. In this paper, we use Lee's method~\cite{esc}, which is known to
 perform well with reasonable computational cost.

\section{Super-Resolution via Sparse Coding} % - - - - - - - - - - - - - - - - - - - - - - - - - - - - - -
The basic idea of single-frame SR based on sparse coding is proposed in~\cite{freeman1}. We
first explain that idea and techniques for improving the quality of
the reconstructed image. Then, we introduce the multi-frame SR method.

\subsection{Inferring HR Image From Single LR image} % - - - - - - - - - - - - - - - - - - - - - - - - - - - - - -
\begin{figure}[t]
	\begin{center}
		\includegraphics[width=12cm]{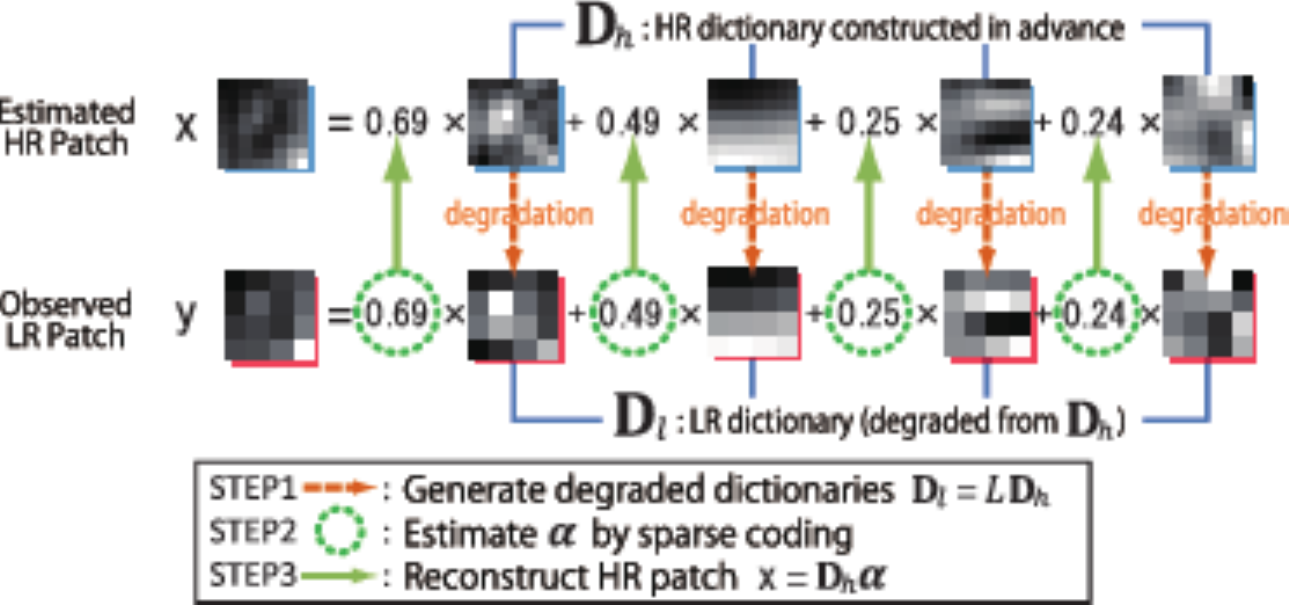}
	\end{center}
 \caption{A framework of super resolution based on sparse coding.\label{fig2}}
\end{figure}

Using HR training images and LR training images, an HR dictionary
 $\mathbf{D}_h \in \mathbb{R}^{q_h \times K}$ and an LR
 dictionary $\mathbf{D}_l \in \mathbb{R}^{q_l \times K}$ are learned before
 performing HR image reconstruction. Assume the HR patch $\bf{x}$ is
 represented by ${\bf{x}} = {\bf{D}}_{h} \boldsymbol \alpha$, and represent the image degradation process $SHW$ by
 $L \in \mathbb{R}^{q_l \times q_h}$. Then, the LR and HR images are connected as
\begin{align}
\mathbf{y} = L{\mathbf{x}} = L\mathbf{D}_h {\boldsymbol \alpha} =
 {\mathbf{D}_{l}} {\boldsymbol \alpha},
\label{eq_scsr}
\end{align}
where ${\mathbf{D}}_{l} = SHW {\mathbf{D}}_{h}$.
In the above equation, we made an assumption, which is the core idea of
sparse coding based SR, that the HR patch shares the same coefficient
with the LR patch in sparse representation. 
For each input LR patch $\mathbf{y}$, a sparse representation
with respect to $\mathbf{D}_{l}$ is found. Then, the corresponding HR bases 
in $\mathbf{D}_{h}$ is combined according to these coefficients to
generate the output HR patch $\bf x$ (Fig.~\ref{fig2}).

\subsection{Reducing Artifact at Patch Boundaries} % - - - - - - - - - - - - - - - - - - - - - - - - - - - - - -
\label{sec:LocalConsistency}
In patch-based approaches, a sparse representation is obtained first for
 an LR patch then for the corresponding HR patch. Obtaining sparse
representation individually for each local patch does not guarantee the
compatibility between adjacent patches, and we often suffer from block
artifacts at the patch boundaries.
To reduce these block artifacts, overlapping patches are often used.
In~\cite{freeman1} and \cite{yang1}, an additional optimization problem is solved after
patch-wise SR to improve the consistency of adjacent patches.
In our proposed method, for the sake of computational efficiency, we
adopt a simple weighted averaging method with 2-dimensional hanning
window for combining adjacent HR patches.

\subsection{Maintaining Global Consistency by Back-Projection}
\label{sec:GlobalConsistency}
Sparse coding of the form Eq.\,\eqref{eq:l1SC} does not demand exact equality
$\mathbf{D}\boldsymbol \alpha = \mathbf{y}$. Because of this and also
because of noise and patch-wise processing, the fundamental
assumption denoted by Eq.\,\eqref{single_dgp} for the target image $\mathbf{Y}_{t}$ and $\mathbf{X}$ may not be satisfied after
patch-wise SR. 
Hence, the back-projection~\cite{bp} is performed for maintaining the
global consistency~\cite{yang1,yang2,wang1}.
Let $\mathbf{X}_{0}$ be the obtained high-resolution
image. To minimize the discrepancy between $\mathbf{X}_{0}$ and the HR
image under the model~\eqref{single_dgp}, we solve the following
minimization problem
\begin{align}
\label{eq:GCObj}
\underset{X }{\text{minimize}}
 \| SH\mathbf{X} - \mathbf{Y} \|_2^2 + c\|\mathbf{X}-\mathbf{X}_0\|_2^2
\end{align}
by gradient descent, where $c > 0$ is a trade-off
parameter. The updating formula of the gradient method is explicitly
written by
\begin{align}
\label{eq:GC}
\mathbf{X}_{t+1} = \mathbf{X}_t - \nu[H^\mathrm{T}S^\mathrm{T}(SH\mathbf{X}_t-\mathbf{Y})+c(\mathbf{X}-\mathbf{X}_0)],
\end{align}
where $\nu > 0$ is a step size parameter.

\subsection{Multi-frame Super Resolution by Sparse Coding} % - - - - - -
								% - - - - - - - - - - - - - - - - - - -
								% - - - - -
In multi-frame scenario, each LR image is
assumed to be an outcome of different degradation of the same HR image.
Multi-frame SR is expected to achieve better results using multiple LR
images, though, extraction and integration of useful information 
in LR images are not a trivial task. 

In images observation model~\eqref{eq:SRmodel}, we assume that  $H_{j} = H$ and $S_{j} =S,\; \forall j \in \{
1,\dots,N \}$ in Eq.\,\eqref{eq:SRmodel} for the sake of simplicity.
We also assume $\bm{\varepsilon}_{j}$ for all $j$ follows the same
Gaussian distribution. From observed LR images $\mathbf{Y}_{j},j=1,\dots,N$, patches $\mathbf{y}_{j},j=1,\dots,N$ 
with different sizes are extracted by clipping operators $C_{j} \in \mathbb{R}^{q_{lj} \times p_l}, \; j = 1,\cdots, N$, which 
will be defined later.

As is the case with single-frame SR, let $\mathbf{X}$ denote the HR
image to be reconstructed, $\mathbf{x}$ be a patch extracted from
$\mathbf{X}$, 
$\mathbf{Y}_1,\dots,\mathbf{Y}_N$ be LR images and 
$\mathbf{y}_1,\dots,\mathbf{y}_N$ be patches of LR images
corresponding to $\mathbf{x}$.
For realizing multi-frame SR, we solve the following optimization problem:
\begin{align}
& \underset{ {\boldsymbol \alpha} }{\text{minimize}} \|\Tilde{\mathbf{D}}_l{\boldsymbol \alpha -
 \Tilde{\mathbf{y}}}\|_2^2 + \eta \| {\boldsymbol \alpha}\|_1
 \label{eq:propObj} \\
&\mathrm{with} \hspace{10pt}  \Tilde{\mathbf{D}}_l = \begin{bmatrix}
													 \mathbf{D}_{l1} \\
													 \vdots \\
													 \mathbf{D}_{lN}
													 \end{bmatrix}
															   \hspace{10pt}
												 \mathrm{and}
									\hspace{10pt} \Tilde{\mathbf{y}} =
 \begin{bmatrix} \mathbf{y}_1 \\ \vdots \\ \mathbf{y}_N \end{bmatrix}.
\end{align}
In our framework of multi-frame SR, we first select a target image from
a set of observed LR images, and construct the HR image of the target
image using other auxiliary LR images. 
 Then,
$\mathbf{y}_1$ and $\mathbf{D}_{l1}$ can be extracted and
constructed in the same manner as single-frame SR. We have to
extract and construct other $N-1$ patches and dictionaries.

In~\cite{ueda2008,wang1}, to alleviate the inaccuracy of integer-pixel accuracy block matching,
referring the technique in the literature of
noise-reduction~{\cite{nlmfilter}}, they used similarities calculated
in the block matching procedure as weights for integrating the
coefficients obtained by sparse coding of each auxiliary patches $\mathbf{y}_{j}, \; j=2,\dots,N$.
 Their method does not involve explicit sub-pixel accuracy block matching. By adopting sub-pixel accuracy block matching,
 we can expect higher accuracy in the HR image reconstruction.

\section{Proposed Method} % - - - - - - - - - - - - - - - - - - - - - - - - - - - - - -
\label{sec_prop} % - - - - - - - - - - - - - - - - - - - - - - - - - - - - - -
An important characteristic of the proposed multi-frame super resolution
method based on sparse coding is in considering the 
sub-pixel accuracy image registration. In the following three
subsections, we explain the basic idea and the concrete procedure for
the proposed method\footnote{A simple implementation of the proposed
method will be available from author's website or journal's support website.}. 

\subsection{General Description} % - - - - - - - - - - - - - - - - - - - - - - - - - - - - - -
\begin{figure}[t]
	\begin{center}
		\includegraphics[width=10cm]{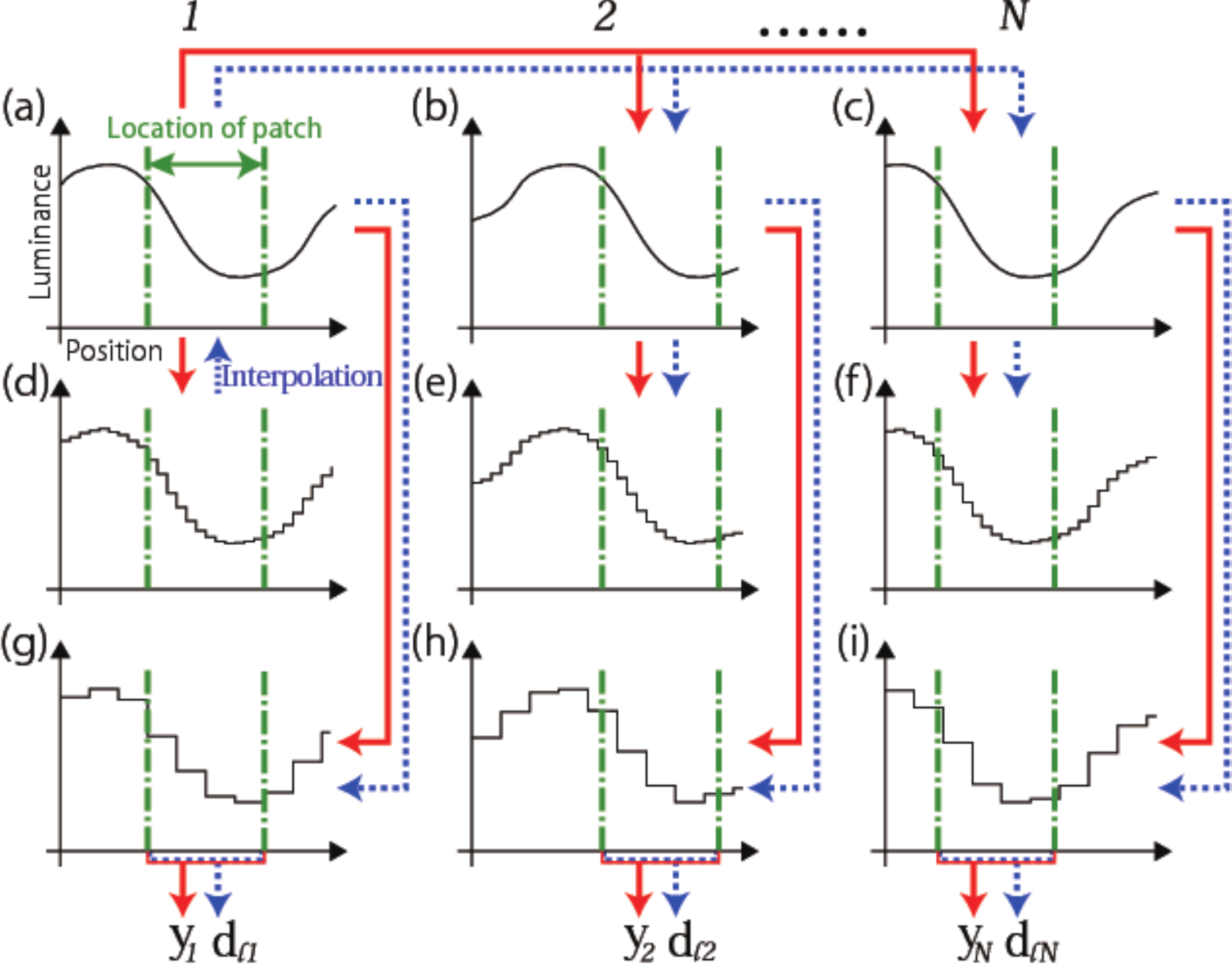}
	\end{center}
 \caption{Sampling considering sub-pixel shift.}
	\label{proc_img}
\end{figure}
We propose to generate a stacked dictionary $\tilde{\mathbf{D}}_l \in
\mathbb{R}^{q_J \times K}$ and stacked observation vectors
$\tilde{\mathbf{y}} \in \mathbb{R}^{q_J}$, taking into account the
displacements between LR images in sub-pixel level. 
To encourage the intuitive understanding, in Fig.~\ref{proc_img}, we
show a conceptual diagram in which images are replaced with
one-dimensional signals. In this figure, horizontal axis and vertical
axis represent location in the image and luminance value, respectively.

In Fig.~\ref{proc_img}, bottom three images (g), (h), and (i) correspond
to the given $N$ observed LR images, and their HR counterparts are shown
as (d), (e), and (f).  We suppose these LR and HR images are sampled
from continuous images (a), (b), and (c), that is, Eqs.~\eqref{single_X} and \eqref{single_Y}
show relationships between (a) and (d), and (a) and (g),
respectively. On the other hand, Eq.\,\eqref{single_dgp} shows relationship
between (d) and (g). 
Given the shift information, continuous images (a), (b), and (c) are
inter-commutable. On the other hand, because of discretization, it is
not possible to convert one discrete HR image to another precisely. In
Fig.~\ref{proc_img}, solid arrows depict the image degradation processes,
dotted arrows depict atom generation processes, and dashed lines depict
positional relationships in sub-pixel accuracy. 

Observed LR images $\mathbf{Y}_{1}, \dots, \mathbf{Y}_{N}$ are assumed
to be down-sampled and blurred versions of continuous images, and they
share the same degradation process except warping process. 
 Let patches $\mathbf{y}_{1},\dots, \mathbf{y}_{N}$ be obtained by
clipping LR images $\mathbf{Y}_{1},\dots,\mathbf{Y}_{N}$ taking into account 
the sub-pixel accuracy shifts estimated by the block matching procedure
introduced in section~\ref{sec_bm}. 
After clipping LR patches $\mathbf{y}_{1},\dots,\mathbf{y}_{N}$, we
generate LR atoms from the HR atoms learned by a dictionary learning
algorithm. Suppose the signal shown in Fig.~\ref{proc_img} (d) is one of
the HR atoms. We can approximate the continuous atom (a) by
interpolation of the atom in (d), and, by parallel shift for continuous
images, we can also approximate warped continuous atoms (b) and
(c). Then, we obtain warped LR atoms (h) and (i) from (b) and (c). In
actual implementation, as described in the next section, we will
calculate warped HR atoms (e) and (f) directly from (d) by
interpolation, and obtain (h) and (i) by applying blur and
down-sampling operators. 

\subsection{Implementation} % - - - - - - - - - - - - - - - - - - - - - - - - - - - - - -
\label{sec_imp}
In this section, we present a concrete procedure for generating the
stacked observation vector $\Tilde{\mathbf{y}}$ and stacked LR
dictionary $\tilde{\mathbf{D}}_{l}$, which are the core
components of the proposed SR method.

For generating the stacked observation vector, patches
$\mathbf{y}_{1},\dots,\mathbf{y}_{N}$ are extracted from auxiliary LR
images by clipping operators $C_j \in \mathbb{R}^{q_{lj} \times p_l}\; j
= 1,\cdots, N$. The clipping operators are designed to extract only
 the pixels inside the cutoff line as shown in~Fig.~\ref{makey}.
Then, the clipped images are stacked to construct $\tilde{\mathbf{y}}$ as
\begin{align}
\Tilde{\mathbf{y}} = \begin{bmatrix} \mathbf{y}_1 \\ \vdots \\ \mathbf{y}_N \end{bmatrix}
= \begin{bmatrix} C_1\mathbf{Y}_1 \\ \vdots \\ C_N\mathbf{Y}_N \end{bmatrix}.
\end{align}

We next explain how to generate the stacked LR dictionary
$\tilde{\mathbf{D}}_{l}$. 
As shown in Eq.\,\eqref{eq:propObj}, $\tilde{\mathbf{D}}_{l}$ is composed of
dictionaries $\mathbf{D}_{lj}, \;j=1,\dots,N$ for
$\mathbf{y}_{j},\;j=1,\dots,N$. Each dictionary $\mathbf{D}_{lj}$ is
generated as follows. Remember that we already have a dictionary
$\mathbf{D}_{h}$ for HR images. Each HR atom in $\mathbf{D}_{h}$ is
embedded into the HR patch region, then degenerated according to the
image degradation processes $S$ and $H$ to obtain LR atoms  
(see Fig.~\ref{proc_img} (d) and (g), and Fig.~\ref{makeDl}: transition
from HR space to LR space). 
Parallel translation in HR space is performed using bi-linear
interpolation~\cite{bl}. 
When the luminance value at $(x,y)$ is estimated, 
we define $u_1 = \lfloor x \rfloor$ and $u_2 = \lfloor y \rfloor$,
$v_1=x-u_1$, $v_2=x-u_2$. Then, we obtain the luminance value at $(x,y)$
as 
\begin{equation}
X(x,y) =  \begin{bmatrix} 1 \! - \! v_1 & v_1 \end{bmatrix}\!\!
\begin{bmatrix} X[u_1,u_2] & X[u_1\!+\!1,u_2] \\ X[u_1,u_2\!+\!1] & X[u_1\!+\!1,u_2\!+\!1] \end{bmatrix}\!\!
\begin{bmatrix} 1 \! - \! v_2 \\ v_2 \end{bmatrix}.
\end{equation}
To obtain an operator $W$ for this sub-pixel accuracy translation, 
we calculate the weighted sum of four pixel-level warp matrices. Let
$w_1, w_2 \in \mathbb{N}$ be horizontal and vertical shifts,
respectively, and
$W_{[w_1,w_2]}$ be the warp operator. 
 Then, the operator $W$ is obtained by 
\begin{multline}
W = (1-v_1)(1-v_2)W_{[w_1,w_2]} + (1-v_1)v_2W_{[w_1+1,w_2]}\\
+ v_1 (1-v_2) W_{[w_1,w_2+1]} + v_1 v_2 W_{[w_1+1,w_2+1]}.
\end{multline}

Atoms for LR images are calculated by clipping operation from the
degraded HR atoms. Then, they are stacked to obtain
$\tilde{\mathbf{d}}_{l}$ as shown in Fig.~\ref{makeDl}.
 From every atom in HR dictionary $\mathbf{D}_{h}$, we calculate the
 stacked LR atom
$\tilde{\mathbf{d}}_{l}$ as presented above, and obtain the stacked LR
dictionary $\Tilde{\mathbf{D}}_l$. 
Denoting the embedding operator $R \in \mathbb{R}^{p_h \times q_h}$,
which embeds an HR atom into the location of target HR patch,
$\Tilde{\mathbf{D}_l}$ is represented by
\begin{align}
\Tilde{\mathbf{D}_l} = \begin{bmatrix} \mathbf{D}_{l1} \\ \vdots \\ \mathbf{D}_{lN} \end{bmatrix}
= \begin{bmatrix} C_1 S H W_1 R \mathbf{D}_h \\ \vdots \\ C_N S H W_N R \mathbf{D}_h \end{bmatrix}.
\end{align}

\subsection{Summary of the Proposed Method} % - - - - - - - - - - - - - - - - - - - - - - - - - - - - - -
\begin{figure}[t!]
  \begin{center}
   \includegraphics[width=10cm]{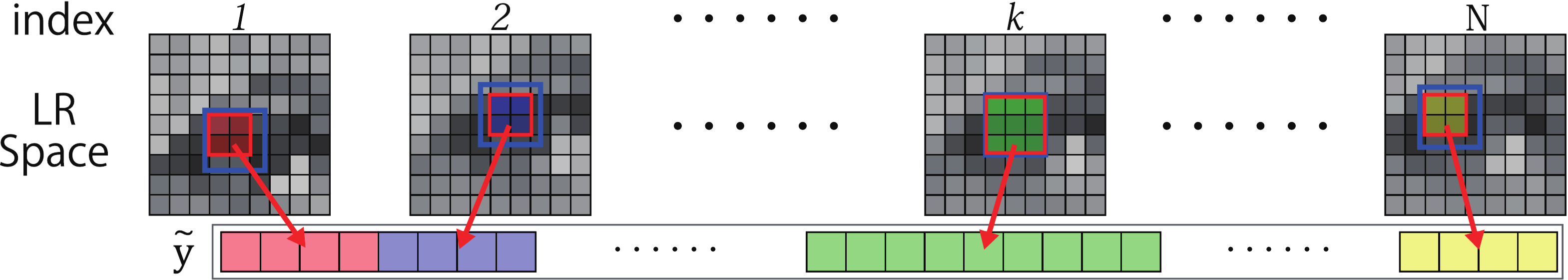}
  \end{center}
  \caption{The procedure of generating stacked LR observation
  $\Tilde{\mathbf{y}}$ by clipping operation}
  \label{makey}
 \end{figure}

\begin{figure}[t!]
  \begin{center}
   \includegraphics[width=10cm]{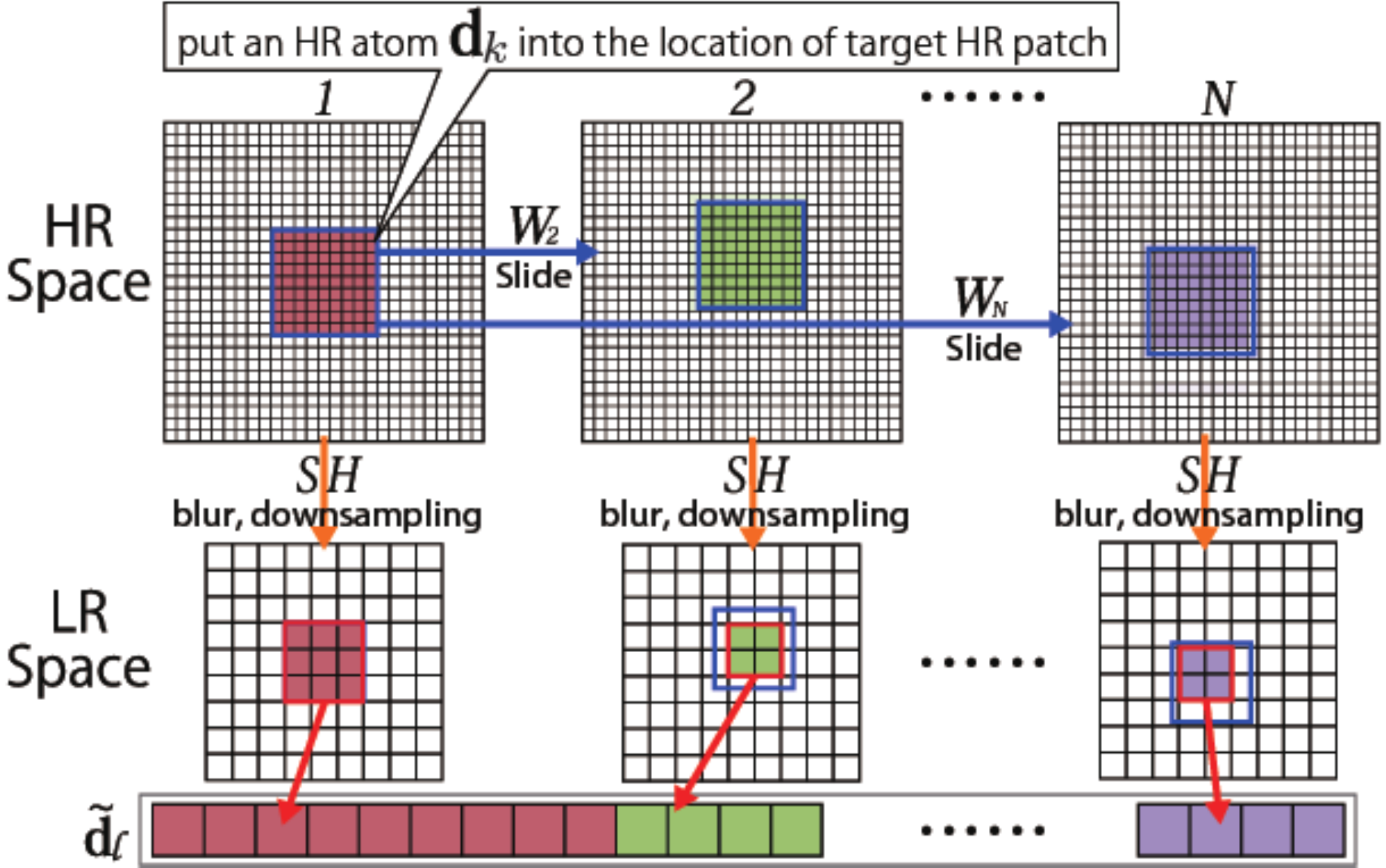}
  \end{center}
  \caption{The procedure of generating stacked LR dictionary $\Tilde{\mathbf{d}}_{lk}$.}
 \label{makeDl}
 \end{figure}

We summarize the proposed method as a concrete procedure in Algorithm~\ref{MfScSR}.
First of all, we arbitrary fix the target image from a
set of $N$ observed images. We let the first image as the target image for
the sake of notational simplicity. The output of the algorithm is the HR
version of this target image $\mathbf{Y}_{1}$. The rest of the algorithm
is composed of two stages, the super-resolution stage, and the
post-processing stage.
It is supposed that an HR dictionary is learned from a set of HR images
beforehand.
In the super-resolution stage, the following procedures are performed
for every small patch $\mathbf{y}_{1}$ extracted from the target image
$\mathbf{Y}_{1}$, and then the tentative HR image $\mathbf{X}_0$ is
obtained by weighted average of the small patches, as described
in~\ref{sec:LocalConsistency}.
By block matching method, relative displacements of LR patches extracted
from LR images $\mathbf{Y}_{j},\; j=2,\dots,N$
are estimated in sub-pixel accuracy. Then, based on this registration
results, the stacked image and stacked
dictionary are generated as shown in Fig.~\ref{makey} and
Fig.~\ref{makeDl}, respectively.
 In an LR image $\mathbf{Y}_{j},\; j=2,\dots,N$, if there are no
block with similarity higher than a predefined threshold value, we do
not use other LR images than $\mathbf{Y}_1$. 
Then, 
 the coefficient $\boldsymbol\alpha$ is estimated for sparse
 representation of $\tilde{\mathbf{y}}_l$ by the dictionary $\tilde{\mathbf{D}}_l$. 
We note that when estimating the coefficient $\bm{\alpha}$ by sparse
coding algorithm, the mean value $m$ of $\tilde{\mathbf{y}}$ is
removed from each element of $\tilde{\mathbf{y}}$, and added to each element of $\mathbf{D}_{h}
\bm{\alpha}$ when reconstructing HR patch $\mathbf{x}$.
Finally, by the HR dictionary $\mathbf{D}_h$ and coefficient
$\boldsymbol\alpha$, the HR patch $\hat{\mathbf{x}}$ is constructed.
In the post-processing stage, the collection of the tentative HR images
$\mathbf{X}_{0}$ is revised by gradient method as stated in~\ref{sec:GlobalConsistency}.
\begin{algorithm}                      
\caption{Multi-frame Super-Resolution via Sparse Representation}      
\label{MfScSR}
\begin{algorithmic}
\STATE {$\mathbf{Input}$: LR Images
 $\mathbf{Y}_1,\mathbf{Y}_2,...,\mathbf{Y}_N$ and HR Dictionary
 $\mathbf{D}_h$. A threshold $\delta \in [0,1]$.}
\STATE{{\{Super-Resolution Stage\}}}
\STATE{Select target image (${\bf{Y}}_{1}$ for example)} 
\FOR{each patch $\mathbf{y}$}
\STATE{-[Step 1] Perform block matching and select informative patches
 based on $\delta$}
\STATE{-[Step 2] Make $\tilde{\mathbf{y}}$,$ \tilde{\mathbf{D}}_{l}$ \{cf. section~\ref{sec_bm} and Figs.~\ref{makey},~\ref{makeDl}\}}
\STATE{-[Step 3] Perform sparse coding}
\STATE{$m \leftarrow \mathbf{mean}(\Tilde{\mathbf{y}})$}
\STATE{$\Tilde{\mathbf{y}} \leftarrow \Tilde{\mathbf{y}} - m$}
\STATE{${\boldsymbol \alpha} \leftarrow \mathbf{Sparse Coding}(\Tilde{\mathbf{D}}_{l},\Tilde{\mathbf{y}})$}
\STATE{-[Step 4] Reconstruction}
\STATE{$\mathbf{x} \leftarrow \mathbf{D}_h {\boldsymbol \alpha} + m$}
\ENDFOR
\STATE{Make $\mathbf{X}_0$ from all patches $\mathbf{x}$} \{cf. section~\ref{sec:GlobalConsistency}\}
\STATE{{\{Post-Processing Stage\}}}
\REPEAT
\STATE{$\mathbf{X} \leftarrow \mathbf{X} -
 \nu[H^\mathrm{T}S^\mathrm{T}(SH\mathbf{X} - \mathbf{Y})+c(\mathbf{X}-\mathbf{X}_{0})]$}
\UNTIL{convergence}
\end{algorithmic}
\end{algorithm}

\subsection{Relationship to Other Methods}
Finally, we conclude this section by comparing the proposed method to
existing SR methods based on sparse coding, and elucidate the advantages
of the proposed method. The comparison among various methods is
summarized in Tab.~\ref{hikaku}.

 To the best of the authors' knowledge, there are only one extant
 multi-frame SR method based on sparse
 coding~\cite{wang1}. In~\cite{wang1}, all of the observed LR images are
 used for HR image construction, while in the proposed method, only
 informative LR images for reconstruction are selected for each patch.

Multi-frame SR requires accurate registration of observed LR images. In
our proposed method, the block matching method with sub-pixel
accuracy~\cite{2dest2} is adopted, and the matching result is
efficiently utilized for stacked LR dictionary construction. Using the
similarity scores obtained in this registration step, we select
informative patches to be used for HR image reconstruction.

In sparse coding stage of SR, both the $\ell^0$-norm and the
	  $\ell^1$-norm are possible for the regularization of the
	  coefficients. The proposed method adopt the $\ell^{1}$-norm,
	  because we observed that it is more stable than the
	  $\ell^{0}$-norm for SR tasks. We used the LARS-Lasso algorithm for
	  efficiently solving the sparse coding problem~\eqref{eq:l1SC}.

 For learning the HR dictionary $\mathbf{D}_{h}$, we apply Lee's
	  method~\cite{esc} which is known to be fast and accurate. The LR
	  dictionary is generated by atoms in $\mathbf{D}_{h}$ according to
	  the assumed blur, down-sampling and estimated translation. This
	  strategy of generating an LR dictionary is more faithful to the
	  idea of super-resolution than the joint dictionary learning
	  strategy adopted in~\cite{yang2,wang1}.

 In order to focus on perceptually important high-frequency content of
 the image, Yang et al.~\cite{yang1} proposed to apply four
 differentiation filters to LR images and applied sparse coding for
 these extracted features. Feature extraction may improve the
 reconstruction accuracy, though, to keep the algorithm simple and to
 avoid increasing of computational costs, we did not use the feature extraction filter and obtained satisfiable results.
\setlength{\tabcolsep}{0.5mm}
\begin{table*}[htbp]
	\caption{Comparison to extant works on sparse coding based
 SR.\label{hikaku}}
\begin{center}
\scalebox{.65}{
\begin{tabular}{|c|c|c|c|r@{}l|c|c|c|c|} \hline
		& input	& registration	& sparse coding	&
 \multicolumn{2}{c|}{dictionary learning}	& feature extraction &
 \shortstack{Local consistency at\\ patch boundary}	&
	 \shortstack{post-processing} \\ \hline
% Proposed
proposed 	&	multiple	& \shortstack{sub-pixel\\block-matching}
 & \shortstack{$\ell^1\text{norm SC}$\\{\small (LARS-Lasso)}} &
	 \shortstack[r]{$\mathbf{D}_h$\\$\mathbf{D}_l$} & \shortstack[l]{:
	 Lee's method\\: $L(\mathbf{D}_h)$} & -- & weighted averaging & back-projection\\ \hline
% Wang
\shortstack{P.~Wang et al.\\(2011) {\cite{wang1}}} & multiple &
		 \shortstack{pixel\\block-matching} &
	 \shortstack{$\ell^1\text{norm SC}$} &
		 \multicolumn{2}{l|}{\shortstack[l]{$\mathbf{D}_h$ and
		 $\mathbf{D}_l$: Lee's method \\ (\small{Joint Dictionary
 Learning})}} & \shortstack{differentiation\\filter} &
	 \shortstack{weighted SC} & back-projection \\ \hline
% Yang1
\shortstack{J.~Yang et al.\\(2008) {\cite{yang1}}} & single & -- &
			 \shortstack{$\ell^1\text{norm SC}$} &
				 \shortstack[r]{$\mathbf{D}_h$\\$\mathbf{D}_l$} &
	 \shortstack[l]{: random sampling \\: $L(\mathbf{D}_h)$} &
	 \shortstack{differentiation\\filter} & \shortstack{weighted SC} &
		 back-projection \\ \hline
\shortstack{J.~Yang, et al.\\(2010) {\cite{yang2}}} & single & -- &
			 \shortstack{$\ell^1\text{norm SC}$} &
				 \multicolumn{2}{l|}{\shortstack[l]{$\mathbf{D}_h$
				 {\small \&} $\mathbf{D}_l$: Lee's method \\
 (\small{Joint Dictionary Learning})}} & \shortstack{differentiation\\filter} &
		 \shortstack{weighted SC} & back-projection \\ \hline
% Yang2
\shortstack{R.~Zeyde et al.\\(2010) {\cite{elad1}}} &
	 single & -- & \shortstack{$\ell^0\text{norm SC}$\\(OMP)} &
	 \shortstack[r]{$\mathbf{D}_h$\\ \\$\mathbf{D}_l$} &
	 \shortstack[l]{: find the best\\dictionary with $\mathbf{D}_l$\\:
 K-SVD} & \shortstack{differentiation\\filter} & \shortstack{averaging\\
 {\huge(}\shortstack{overlap is taken account\\in
 dictionary learning}{\huge)}} & -- \\ \hline
% Min
\shortstack{L.~Min et al.\\(2010) {\cite{sksvdsr}}} & single & -- &
			 \shortstack{$\ell^1\text{norm SC}$} &
				 \multicolumn{2}{c|}{\shortstack[l]{$\mathbf{D}_h${\small
				 \&}$\mathbf{D}_l$: Sparse K-SVD{\cite{sksvd}} \\
 (\small{Joint Dictionary Learning})}} & \shortstack{differentiation\\filter} &
		 weighted SC & back-projection \\ \hline
% Dong
\shortstack{W.~Dong et al.\\(2011) {\cite{asdssr}}} & single & --
		 & {\small\shortstack{Iterated Shrinkage\\Algorithm for\\ASDS
 with AReg}} & \multicolumn{2}{c|}{\shortstack[l]{Sub-dictionaries for
	 ASDS:\\K-means and PCA}} & \shortstack{high-pass filter} &
		 averaging & -- \\ \hline
% Lu
\shortstack{X.~Lu et al.\\(2012) {\cite{gccssr}}} & single & -- &
			 {\small\shortstack{Geometry\\Constrained\\$\ell^1\text{norm
 SC}$}} & \shortstack[r]{$\mathbf{D}_h$\\ \\$\mathbf{D}_l$} &
	 \shortstack[l]{: find the best\\dictionary with $\mathbf{D}_l$\\:
 Geometry Dictionary} & -- & averaging & -- \\ \hline
\end{tabular}
}
\end{center}
\end{table*}

\section{Experimental Results}
In this section, we first demonstrate the super-resolution results
obtained by the proposed method on some sets of 
still images, and some sets of images from movies. 

In our experiments, we magnify the input LR images by a factor of $3$
for all cases. We used $5$ LR images to estimate an HR images for
multi-frame SR methods.
\subsection{Application to Still Images}
In the experiments with still images, we generated artificially
degraded LR images according to the generative model in Eq.\,\eqref{eq:SRmodel}.
As described in section II, we suppose observed LR images are generated
from an HR image through parallel translations, blurring, down-sampling, and
addition of noises. 
The parallel translations are imitated by shifting
the HR image. The degree of vertical and horizontal shifts are randomly
sampled from a uniform distribution in $[-5,5]$. 
The blurring operation $H$ is realized by convolution of $9 \times
9$-pixel Gaussian filter with the standard deviation $\sigma_h = 1$. 
The blurred and shifted images are then down-sampled by the factor
$3$. Finally, noises sampled from $\mathcal{N}(0,\sigma_{n}=\sqrt{2})$
are added to generate LR observation images. We note that by this noise
addition, SNRs of resulting $5$ images are about $37 \sim 40$[dB].

We randomly selected one target LR image, and estimated relative displacements from
the target to other four images by sub-pixel accuracy
block matching. In our experiments, both the intensity of the blur and
the noise are supposed to be given. We summarize the experimental
settings in table~\ref{impset_img_kansoku}.

\begin{table}[htbp]
	\caption{Experimental settings (for still images)\label{impset_img_kansoku}}
\begin{center}
 	\begin{tabular}{c|c} \hline
	 parameters & values \\ \hline
		number of LR images & 5 \\
	 parallel translation & Unif$(-5,5)$ [px]\\
	 blur & $9 \times 9$ Gaussian filter ($\sigma_h = 1$) \\
	 noise& Gaussian ($\sigma_n = \sqrt{2}$) \\ \hline
	\end{tabular}
\end{center}
\end{table}

We compare the proposed method to four conventional methods.
The first method is bi-cubic interpolation. This method is simple
and regarded as a baseline for SR. The second method is the one proposed
in~\cite{yang1}, which is a
single-frame SR method based on joint dictionary learning. We refer to
this method as SF-JDL (Single Frame super resolution based on Joint
Dictionary Learning). The third method is the one proposed
in~\cite{elad1}, which is based on the $\ell^{0}$ sparse coding and we refer to
this method as SF-L0 (Single Frame super resolution based on the $\ell^0$
sparse coding).
The fourth method is the one proposed in~\cite{wang1}, which is a
multi-frame SR method based on joint dictionary learning. We refer to
this method as MF-JDL (Multi-Frame super resolution based on Joint
Dictionary Learning). 
As for MF-JDL, we used $5$ LR images including a target image for
reconstructing HR images from them.
As for bi-cubic method, SF-JDL and SF-L0, only the target LR image is
used for HR image reconstruction.
As for the relative displacements of LR images in the proposed methods,
we tried two settings. The one is estimating the displacements
by sub-pixel accuracy block matching method, and the other is assuming the ground-truth
displacements are given. The former is referred to as
proposed(1), while the later is refereed to as proposed(2) in this section.

For quantitative comparison of SR methods, we use the Peak Signal to
Noise Ratio (PSNR) defined as
\begin{align}
{\rm{PSNR}}\text{[dB]} = 10 \log_{10} \frac{255^2}{{\rm{MSE}}},
\end{align}
where MSE is the mean squared error between the original HR image and
the estimated HR image, and the higher PSNR indicates the better SR
performance.
We note that SF-L0 is based on the second order differential of images,
and it can not reconstruct three pixels from the edge (rim) of the
image. So, we calculate the PSNR discarding three pixels from the edge
of the reconstructed HR images, and this evaluation may work advantageous
to SF-L0.

There are seven parameters in the proposed method:
\begin{enumerate}
 \item the strength of sparseness constraint $\eta$ in sparse coding 
 \item the number of atoms $m$ in the dictionary used in sparse coding
 \item the size of patch $q_h$
 \item the size of overlapping of patches
 \item the threshold $\delta$ to judge whether a LR patch is used or not based
	   on similarity score obtained by block matching
 \item the balancing parameter $c$ for global consistency~\eqref{eq:GCObj}
 \item the step size $\nu$ for back-projection~\eqref{eq:GC}
\end{enumerate}

According to our preliminary experimental results, we set the values of
these parameters as shown in table~\ref{tab:impset_img_par}. We note that
the essentially new parameter in our method is $\delta$. For other SR
methods compared in this paper, other six parameters are also optimized
according to our preliminary experiments.
In this experiment, we set the threshold $\delta$ to zero and use all
the $5$ LR images for HR image reconstruction.
\begin{table}[htbp]
	\caption{Parameter settings for the proposed method (still image)\label{tab:impset_img_par}}
\begin{center}
 	\begin{tabular}{c|c} \hline
		parameters & values \\ \hline
	 $\eta$: penalty for the $\ell^{1}$-norm in~\eqref{eq:propObj} & 0.05 \\
	 $m$: number of basis & 512 \\
	 $q_{h}$: size of HR patch & $15\times15=125$ \\
	 size of LR patch overlapping & $3$ \\
	 $\delta$: threshold of block matching score & $0.0$ \\
	 $c$: trade-off in~\eqref{eq:GC} & $0.0001$ \\
	 $\nu$: step size in~\eqref{eq:GC} & $0.001$ \\ \hline
	\end{tabular}
\end{center}
\end{table}
We use two different gray-scale images (Lena and Cameraman), and three
color images (Flower, Girl, and Parthenon) for evaluating the
performance of SR methods. When we deal with color images, we first
convert the image into YCbCr format, then apply SR methods only to
luminance channel (Y). Values of other channels Cb and Cr are simply
expanded by bi-cubic interpolation. 

\begin{figure*}[htbp]
  \begin{center}
    \begin{tabular}{cccc}
      % [1]Input
      \begin{minipage}{3.5cm}
        \begin{center}
          \includegraphics[clip, width=3.5cm]{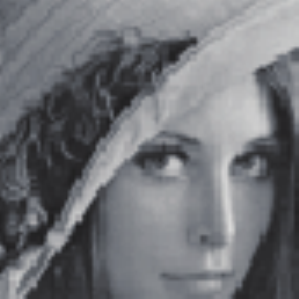}
        \end{center}
      \end{minipage}
	&
      %  [2]Bicubic
      \begin{minipage}{3.5cm}
        \begin{center}
          \includegraphics[clip, width=3.5cm]{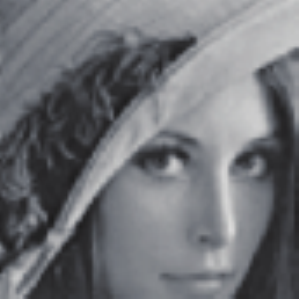}
        \end{center}
      \end{minipage}
	&
      % [3]Yang2010
      \begin{minipage}{3.5cm}
        \begin{center}
          \includegraphics[clip, width=3.5cm]{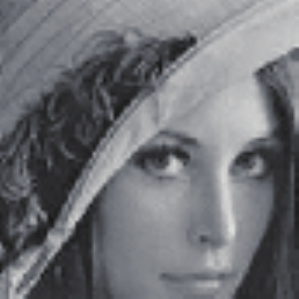}
        \end{center}
      \end{minipage}
	&
      %  [4]zeyde2010
      \begin{minipage}{3.5cm}
        \begin{center}
          \includegraphics[clip, width=3.5cm]{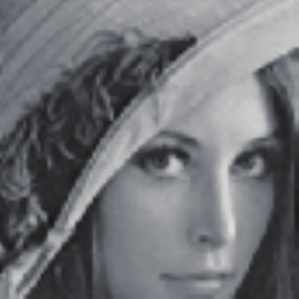}
        \end{center}
      \end{minipage}
	\\
	{\small Input image}
	&
	{\small Bi-cubic}
	&
	{\small SF-JDL}
	&
	{\small SF-L0}
		\\
	%  [4]Wang2012
      \begin{minipage}{3.5cm}
        \begin{center}
          \includegraphics[clip, width=3.5cm]{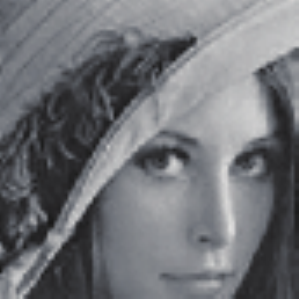}
        \end{center}
      \end{minipage}
	&
      % [5]proposed1
      \begin{minipage}{3.5cm}
        \begin{center}
          \includegraphics[clip, width=3.5cm]{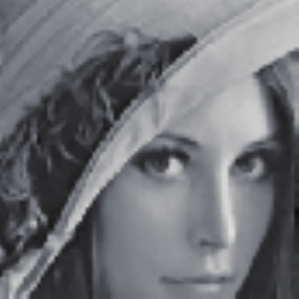}
        \end{center}
      \end{minipage}
	&
      % [6]proposed2
      \begin{minipage}{3.5cm}
        \begin{center}
          \includegraphics[clip, width=3.5cm]{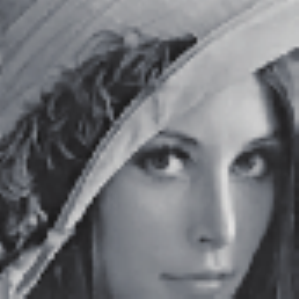}
        \end{center}
      \end{minipage}
	&
      % [7]Original
      \begin{minipage}{3.5cm}
        \begin{center}
          \includegraphics[clip, width=3.5cm]{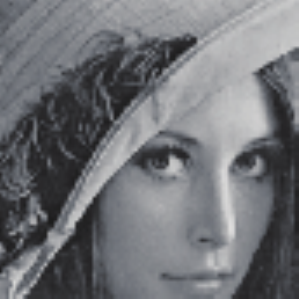}
        \end{center}
      \end{minipage}
	\\
	{\small MF-JDL}
	&
	{\small proposed(1)}
	&
	{\small proposed(2)}
	&
	{\small original image}
    \end{tabular}
    \caption{Images estimated from LR observations (Lena)}
    \label{r_lena}
  \end{center}
\end{figure*}

\begin{figure*}[htbp]
  \begin{center}
    \begin{tabular}{cccc}
      % [1]Input
      \begin{minipage}{3.5cm}
        \begin{center}
          \includegraphics[clip, width=3.5cm]{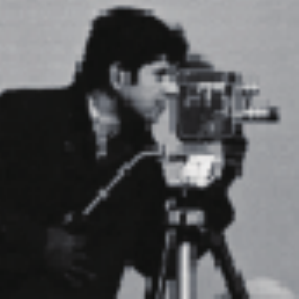}
        \end{center}
      \end{minipage}
	&
      %  [2]Bicubic
      \begin{minipage}{3.5cm}
        \begin{center}
          \includegraphics[clip, width=3.5cm]{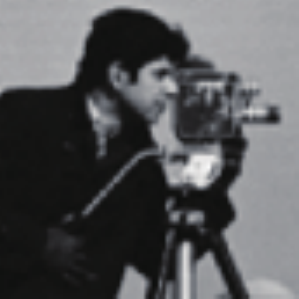}
        \end{center}
      \end{minipage}
	&
      % [3]Yang2010
      \begin{minipage}{3.5cm}
        \begin{center}
          \includegraphics[clip, width=3.5cm]{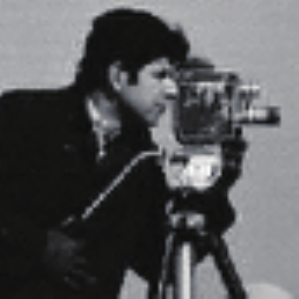}
        \end{center}
      \end{minipage}
	&
      %  [4]Zeyde
      \begin{minipage}{3.5cm}
        \begin{center}
          \includegraphics[clip, width=3.5cm]{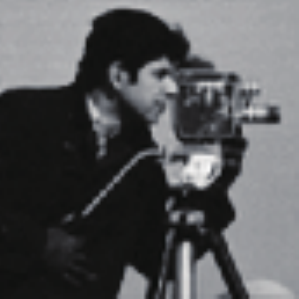}
        \end{center}
      \end{minipage}
	\\
	{\small input}
	&
	{\small Bi-cubic}
	&
	{\small SF-JDL}
	&
	{\small SF-L0}
		\\
	%  [5]Wang2011
	\begin{minipage}{3.5cm}
		\begin{center}
          \includegraphics[clip, width=3.5cm]{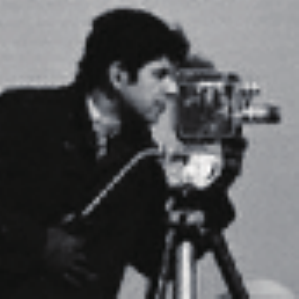}
        \end{center}
	\end{minipage}
	&
      % [6]proposed1
      \begin{minipage}{3.5cm}
        \begin{center}
          \includegraphics[clip, width=3.5cm]{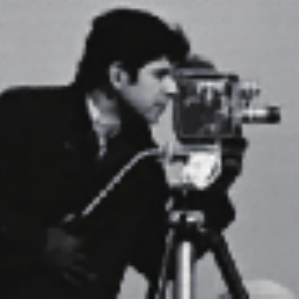}
        \end{center}
      \end{minipage}
	&
      % [7]proposed2
      \begin{minipage}{3.5cm}
        \begin{center}
          \includegraphics[clip, width=3.5cm]{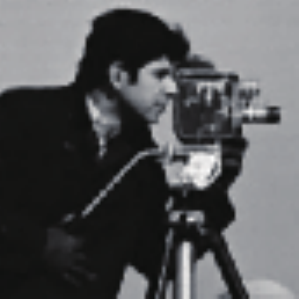}
        \end{center}
      \end{minipage}
	&
      % [8]Original
      \begin{minipage}{3.5cm}
        \begin{center}
          \includegraphics[clip, width=3.5cm]{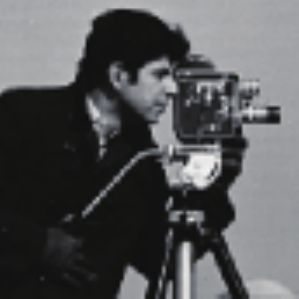}
        \end{center}
      \end{minipage}
		\\
	{\small MF-JDL}
	&
	{\small proposed(1)}
	&
	{\small proposed(2)}
			 &
	{\small original image}
    \end{tabular}
   \caption{Images estimated from LR observations (Cameraman)}
    \label{r_cameraman}
  \end{center}
\end{figure*}

We show the experimental results in Fig.~\ref{r_lena} to
Fig.~\ref{r_cameraman}. We note that we also applied SR methods to the
famous color images (Flower, Girl, and Parthenon), but for the sake of image size, we omit
showing the figures and just show the performace in PSNR measure. In
these figures, the degraded LR image, the
results by bi-cubic interpolation, SF-JDL, SF-L0, MF-JDL, proposed(1),
proposed(2), and the original HR image are shown.
These figures indicate that the proposed method can generate much higher
resolution images with sharp edges.
Not surprisingly, the proposed(2), where the ground-truth displacements
are given, provides the most clear HR images. However, proposed(1), where
the displacements are estimated, provides clearer HR images than that 
obtained by other conventional methods.

We also show PSNR values obtained by various methods in table~\ref{tab:result_psnr}.
For evaluating the PSNR values, we randomly generated $100$ sets of $5$ warp operators and generated 
degraded images by adding random observation noises. From each set of observed images, we randomly choose one target image. 
The only target image is used for single-frame SR, while in multi-frame SR, the remaining $4$ images are used as auxiliary LR images.
The means and standard deviations of PSNR values are calculated using
$100$ SR results by each methods. The best and the second best results
are shown in bold style.
\begin{table*}[ht!]
	\caption{PSNRs of SR methods}
\begin{center}
\scalebox{.8}{
 \label{tab:result_psnr}
	\begin{tabular}{c|ccccccc} \hline
 		Image & Input & Bi-cubic & SF-JDL & SF-L0  & MF-JDL & proposed(1) & proposed(2) \\ \hline	 
		Lena & $26.34\pm0.00 \;$ & $\;28.01\pm 0.01\;$ &
				 $\;28.64\pm 0.02\;$ & $\;29.38\pm 0.01\;$ &
						 $\;29.10 \pm 0.01\;$ & $\;
							 {\mathbf{29.97}}\pm 0.12\;$ & $\;
								 {\mathbf{30.94}}\pm 0.20\;$ \\
		Cameraman & $24.37\pm 0.00 \;$ & $\;26.87\pm 0.01\;$ &
				 $\;27.85\pm 0.02\;$ & $\;29.08\pm 0.01\;$ &
						 $\;28.17\pm 0.02\;$ &
							 $\;{\mathbf{29.72}}\pm 0.41\;$ &
	 $\;{\mathbf{32.24}}\pm 0.35\;$ \\
		Flower & $33.44\pm 0.00\;$ & $\;35.36\pm 0.01\;$ &
				 $\;35.52\pm 0.02\;$ & $\;36.36\pm 0.01\;$ &
						 $\;36.00\pm 0.02\;$ & $\;{\mathbf{36.44}}\pm 0.11\;$ &
	 $\;{\mathbf{36.92}}\pm 0.15\;$ \\
		Girl & $29.81 \pm 0.00\;$ & $\;31.06 \pm 0.00\;$ &
				 $\;31.33\pm 0.01\;$ & $\;31.74 \pm 0.01\;$ &
						 $\;31.55 \pm 0.01\;$ &
							 $\;{\mathbf{31.88}}\pm 0.06\;$ &
	 $\;{\mathbf{32.20}}\pm 0.10\;$ \\
		Parthenon & $23.50\pm 0.00\;$ & $\;24.31\pm 0.00\;$ &
				 $\;24.41\pm 0.00\;$ & $\;24.92\pm 0.00\;$ &
						 $\;24.59\pm 0.00\;$ &
							 $\;{\mathbf{25.36}}\pm 0.10\;$ &
	 $\;{\mathbf{26.39}}\pm 0.15\;$ \\ \hline
	\end{tabular} 
}
\end{center}
\end{table*}
From table~\ref{tab:result_psnr}, the proposed method is shown to
outperform other conventional methods. We note that standard deviations of PSNR values for all of the single-frame SR methods are very low,
because randomness involved in these methods is only due to observation noise. 
The standard deviations of MF-JDL is also low compared to those of the proposed methods. The relatively high standard deviations for the 
proposed methods can be attributed to instability of the sub-pixel accuracy block matching method. 
However, even with relatively high standard deviations, the proposed methods show statistically significant improvements to conventional methods.

\subsection{Application to Motion Pictures}
To see that the proposed method can be applied to more practical
problems, we perform experiments using LR images sequentially captured from movies. 
Images captured from movies are degraded according to the degradation
model specified by table~\ref{impset_img_kansoku} excluding the parallel
translation. 
From five consecutive LR images, the middle (third in the temporal
sequence) image is selected as the target image, and other four are
considered as auxiliary images.

Except for the threshold $\delta$, the same parameters shown in
table~\ref{tab:impset_img_par} are used in this section. The parameter
$\delta$ determines whether an LR image is used for HR image
reconstruction or not based on the matching score~\eqref{eq:simscore}. This parameter is 
particularly important in multi-frame SR applied for LR images captured
from movies. In this experiment, we show experimental results with
$\delta=0.003$ and $\delta=0.001$. When the proposed method is applied to LR
images sequentially captured from movies, an object appears in an LR
image can be hidden behind another object. It is also possible that the
object goes out of the scene. When the threshold is set to $\delta  =
0.003$, only the auxiliary patches which obtain high similarity to the
target patch are used for the HR reconstruction. On the other hand,
 when the threshold is set to $\delta = 0.001$, the algorithm adopts
most of patches cropped from auxiliary LR images.

\begin{figure*}[htbp]
  \begin{center}
    \begin{tabular}{cccc}
      % [1]Input
      \begin{minipage}{3.5cm}
        \begin{center}
          \includegraphics[clip, width=3.5cm]{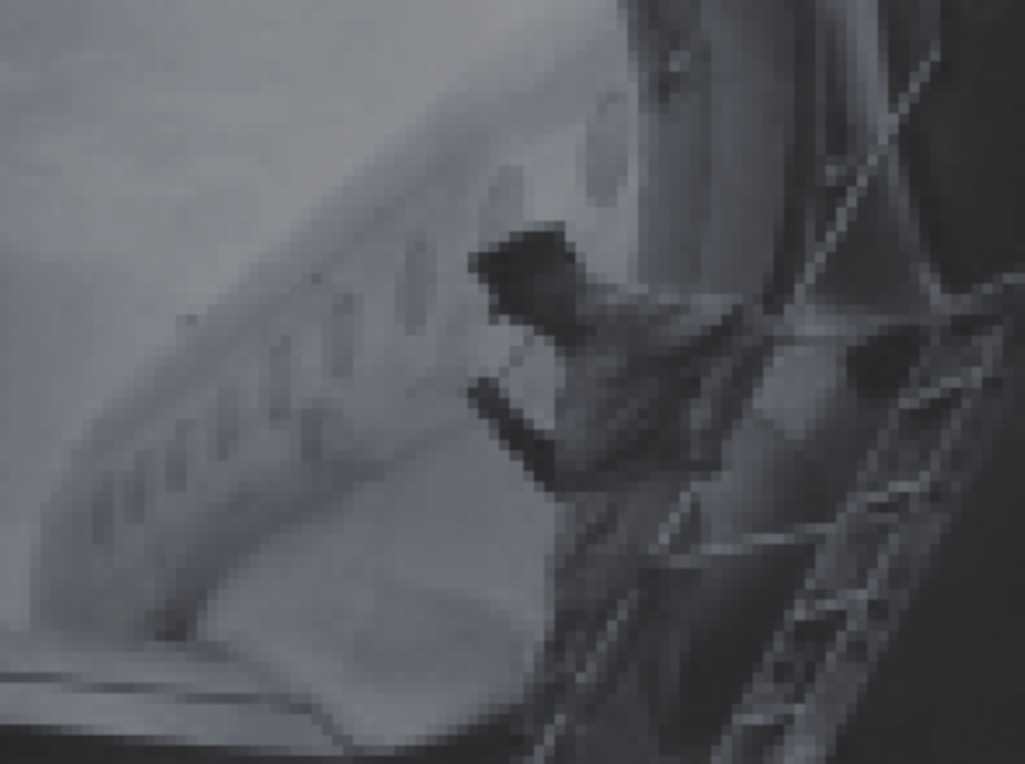}
        \end{center}
      \end{minipage}
	&
      %  [2]Bicubic
      \begin{minipage}{3.5cm}
        \begin{center}
          \includegraphics[clip, width=3.5cm]{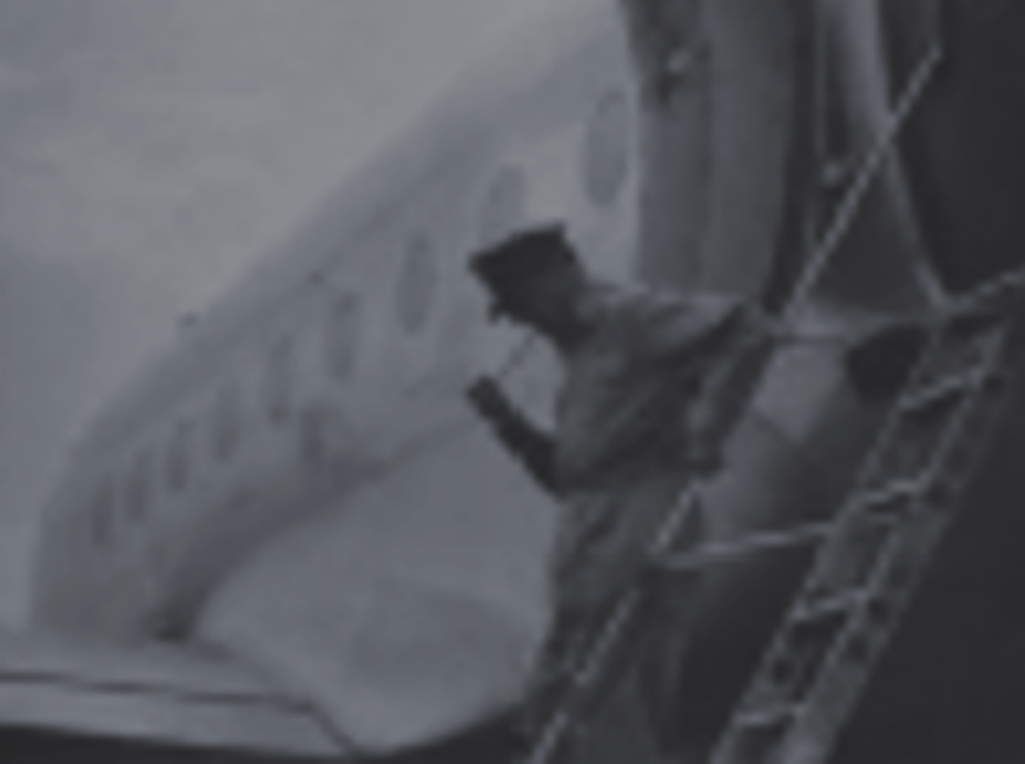}
        \end{center}
      \end{minipage}
	&
      % [3]Yang2010
      \begin{minipage}{3.5cm}
        \begin{center}
          \includegraphics[clip, width=3.5cm]{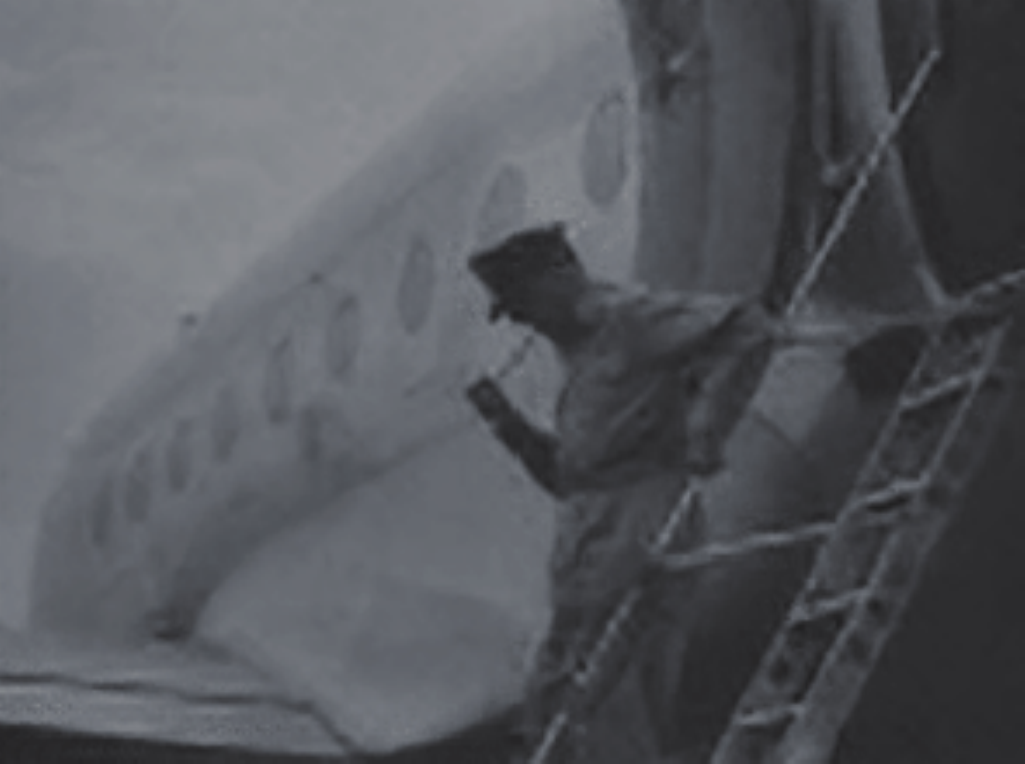}
        \end{center}
      \end{minipage}
	&
      %  [4]Zeyde2010
      \begin{minipage}{3.5cm}
        \begin{center}
          \includegraphics[clip, width=3.5cm]{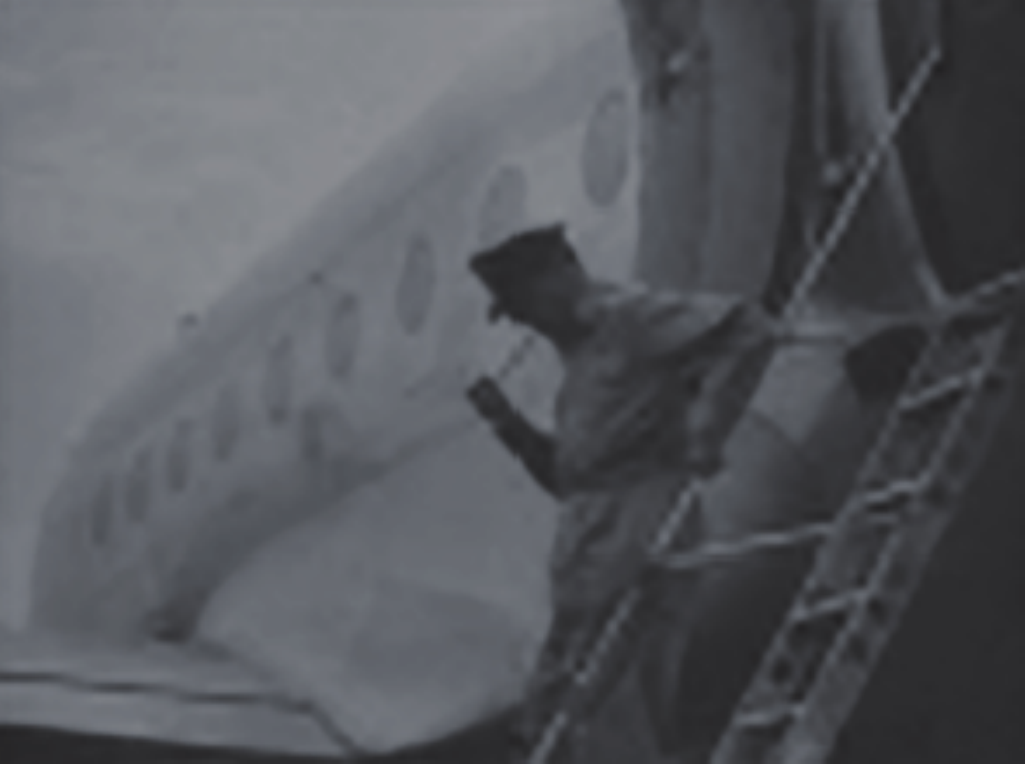}
        \end{center}
      \end{minipage}
	\\
	{\small input}
	&
	{\small Bi-cubic}
	&
	{\small SF-JDL}
	&
	{\small SF-L0}
		\\
	%  [5]Wang2011
      \begin{minipage}{3.5cm}
        \begin{center}
          \includegraphics[clip, width=3.5cm]{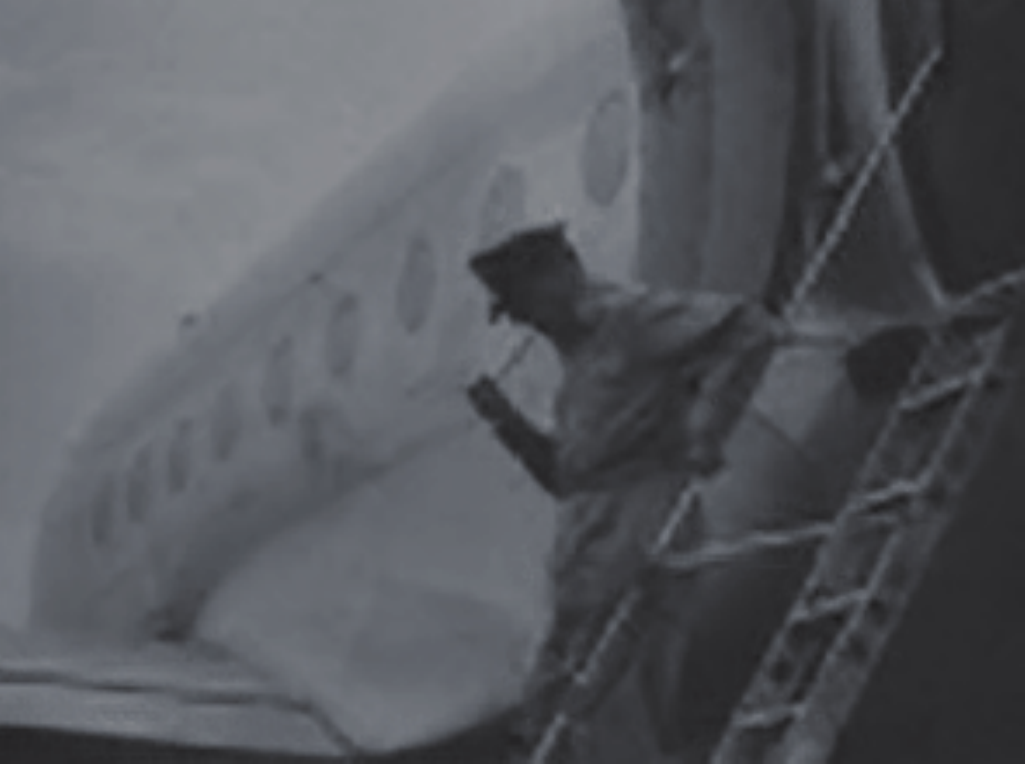}
        \end{center}
      \end{minipage}
	&
      % [6]proposed1
      \begin{minipage}{3.5cm}
        \begin{center}
          \includegraphics[clip, width=3.5cm]{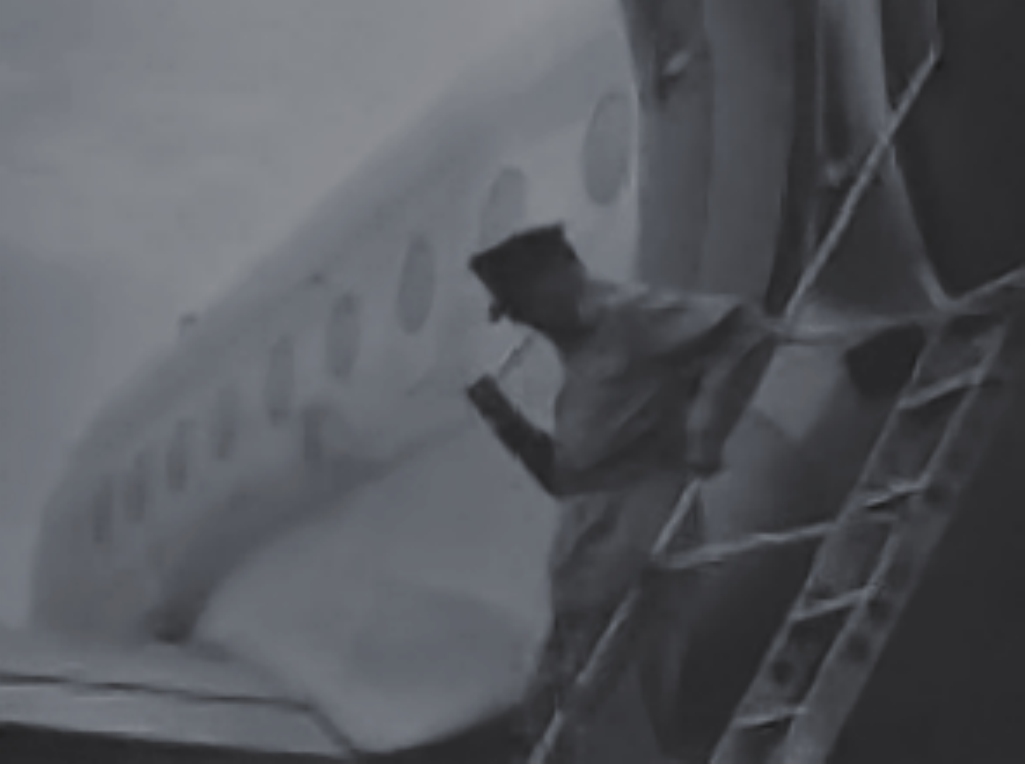}
        \end{center}
      \end{minipage}
	&
      % [7]proposed2
      \begin{minipage}{3.5cm}
        \begin{center}
          \includegraphics[clip, width=3.5cm]{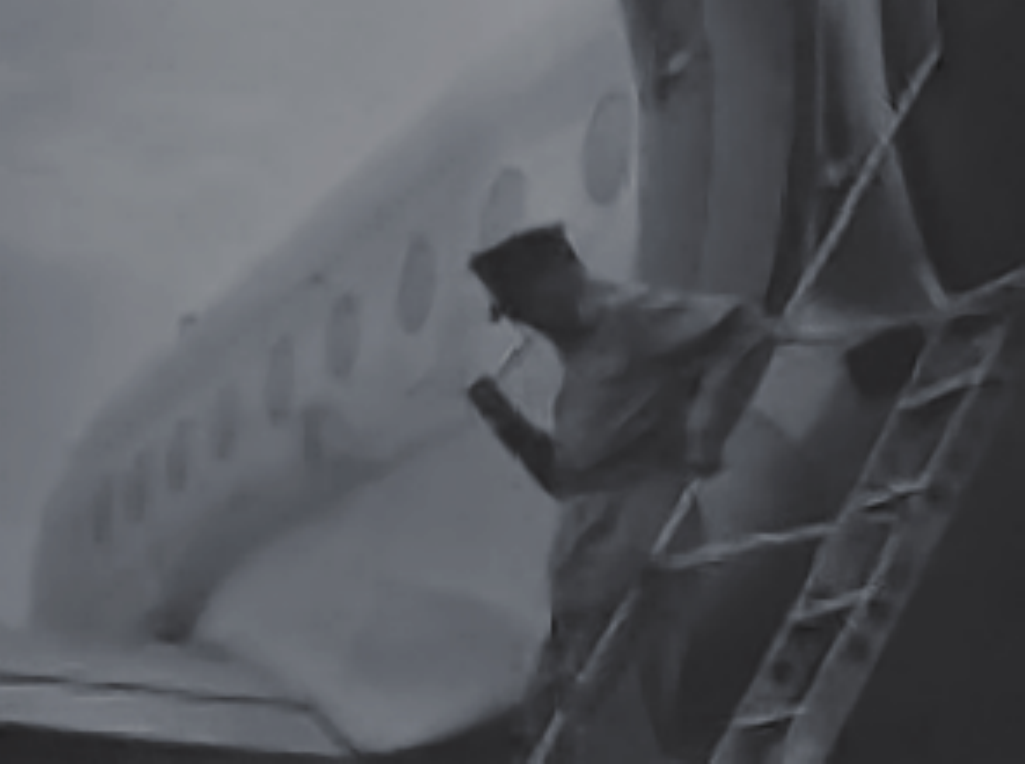}
        \end{center}
      \end{minipage}
	&
      % [8]Original
      \begin{minipage}{3.5cm}
        \begin{center}
          \includegraphics[clip, width=3.5cm]{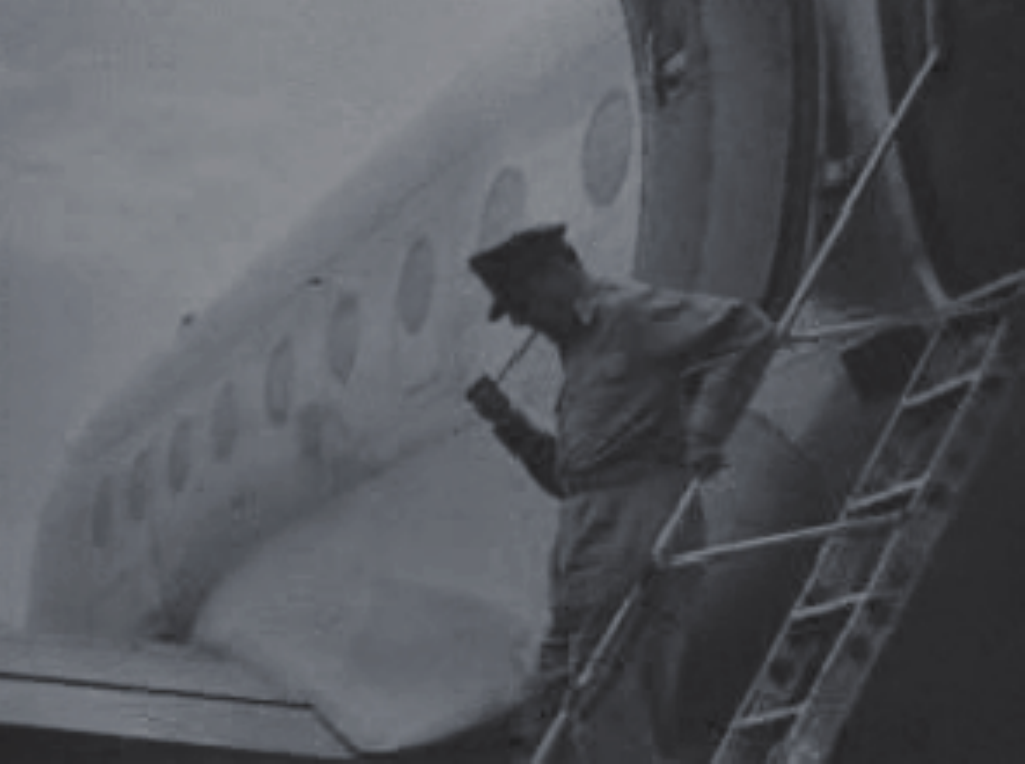}
        \end{center}
      \end{minipage}
		\\
	{\small MF-JDL}
	&
	{\small proposed($\delta=0.003$)}
	&
	{\small proposed($\delta=0.001$)}
	&
	{\small original image}
    \end{tabular}
\caption{Images estimated from LR observations (MacArthur)}
    \label{r_mac}
  \end{center}
\end{figure*}

\begin{figure*}[htbp]
  \begin{center}
    \begin{tabular}{cccc}
      % [1]Input
      \begin{minipage}{3.5cm}
        \begin{center}
          \includegraphics[clip, width=3.5cm]{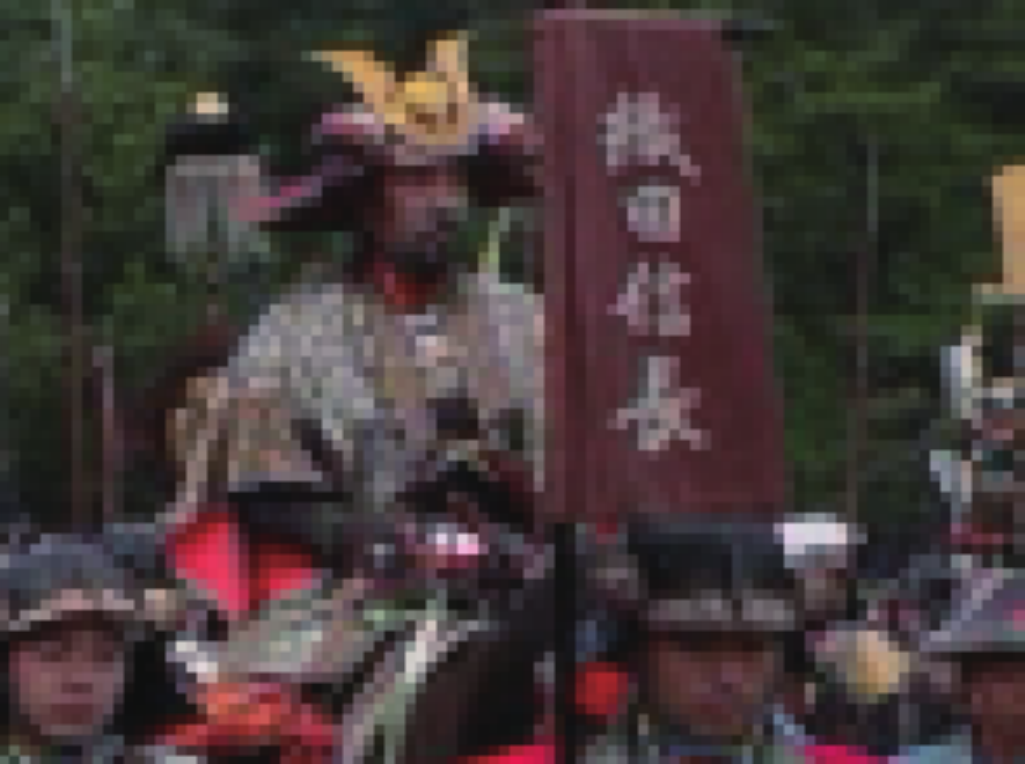}
        \end{center}
      \end{minipage}
	&
      %  [2]Bicubic
      \begin{minipage}{3.5cm}
        \begin{center}
          \includegraphics[clip, width=3.5cm]{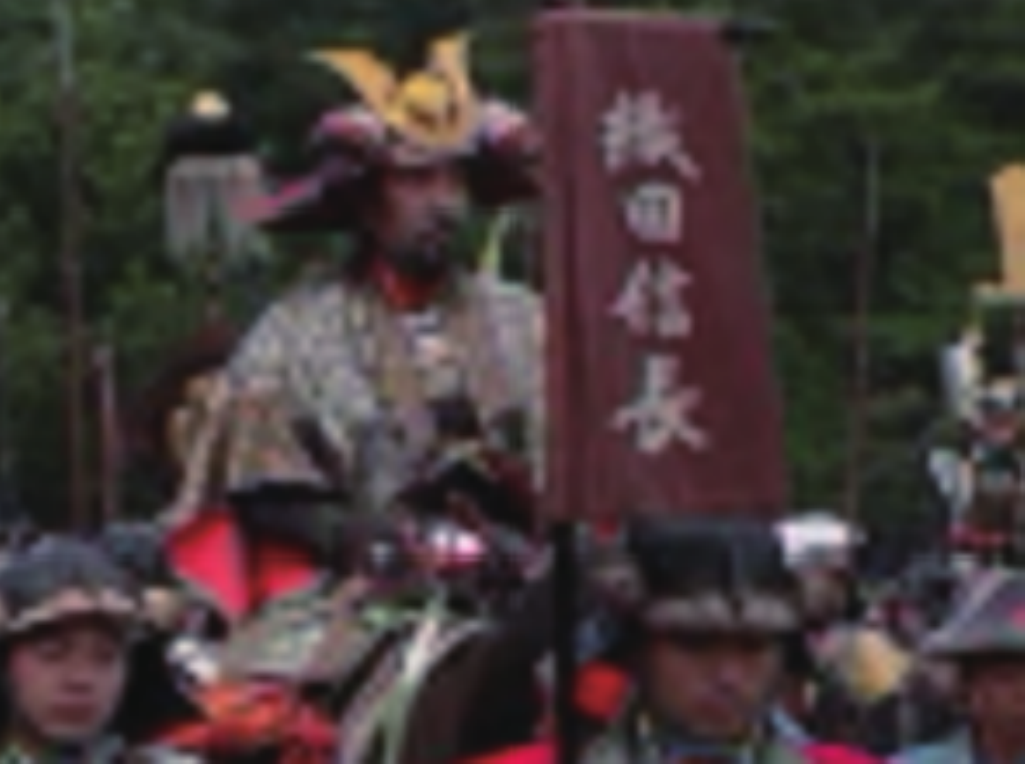}
        \end{center}
      \end{minipage}
	&
      % [3]Yang2010
      \begin{minipage}{3.5cm}
        \begin{center}
          \includegraphics[clip, width=3.5cm]{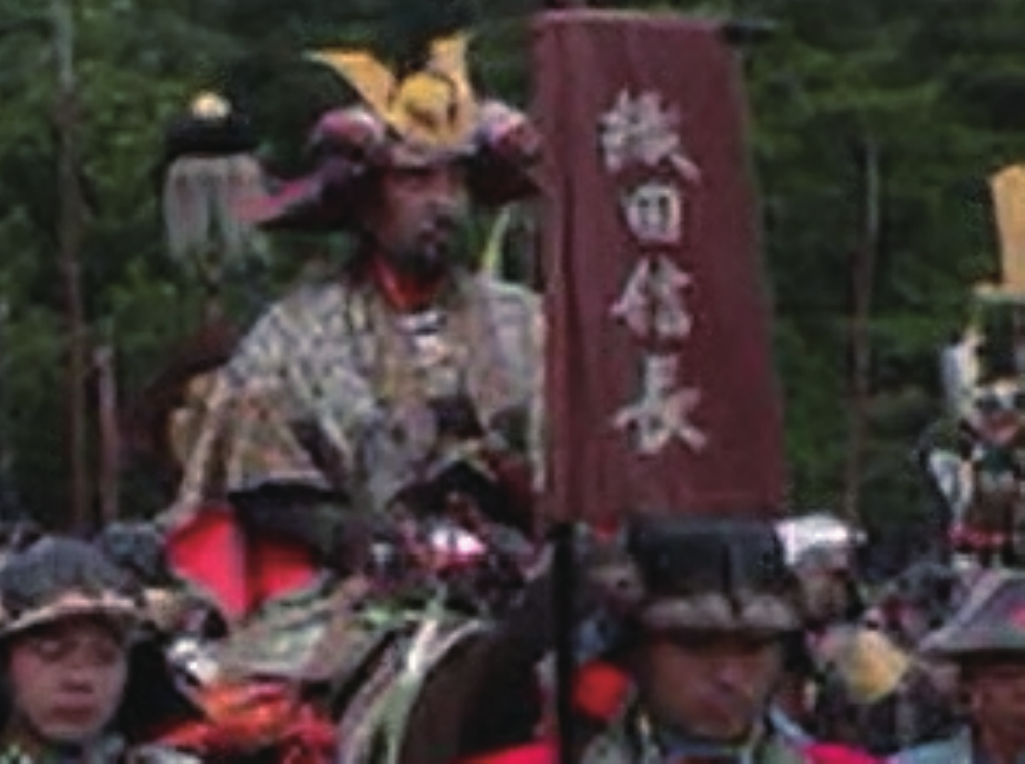}
        \end{center}
      \end{minipage}
	&
      %  [4]Zeyde2010
      \begin{minipage}{3.5cm}
        \begin{center}
          \includegraphics[clip, width=3.5cm]{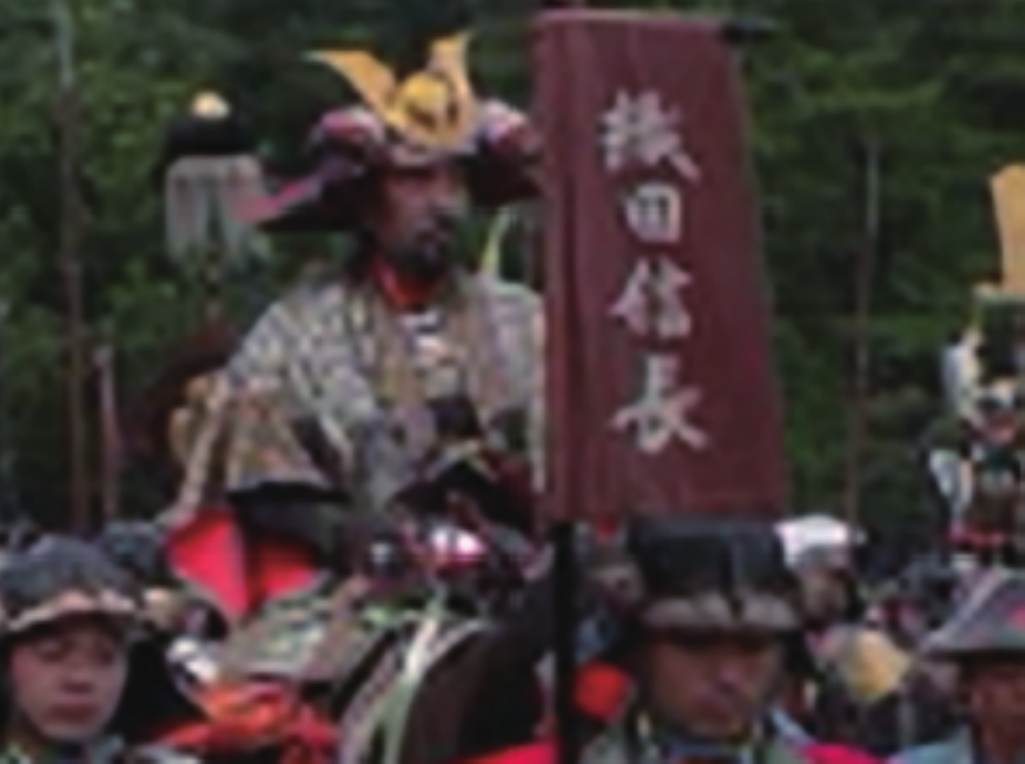}
        \end{center}
      \end{minipage}
	\\
	{\small input}
	&
	{\small Bi-cubic}
	&
	{\small SF-JDL}
	&
	{\small SF-L0}
		\\
	%  [4]Wang2011
      \begin{minipage}{3.5cm}
        \begin{center}
          \includegraphics[clip, width=3.5cm]{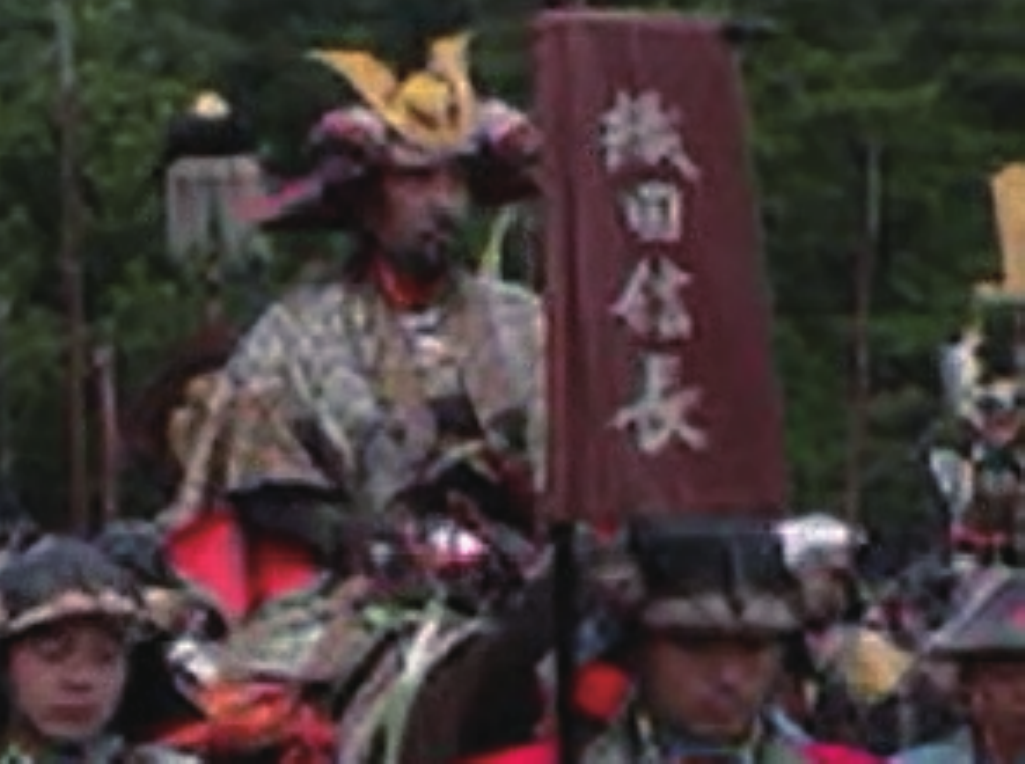}
        \end{center}
      \end{minipage}
	&
      % [5]proposed1
      \begin{minipage}{3.5cm}
        \begin{center}
          \includegraphics[clip, width=3.5cm]{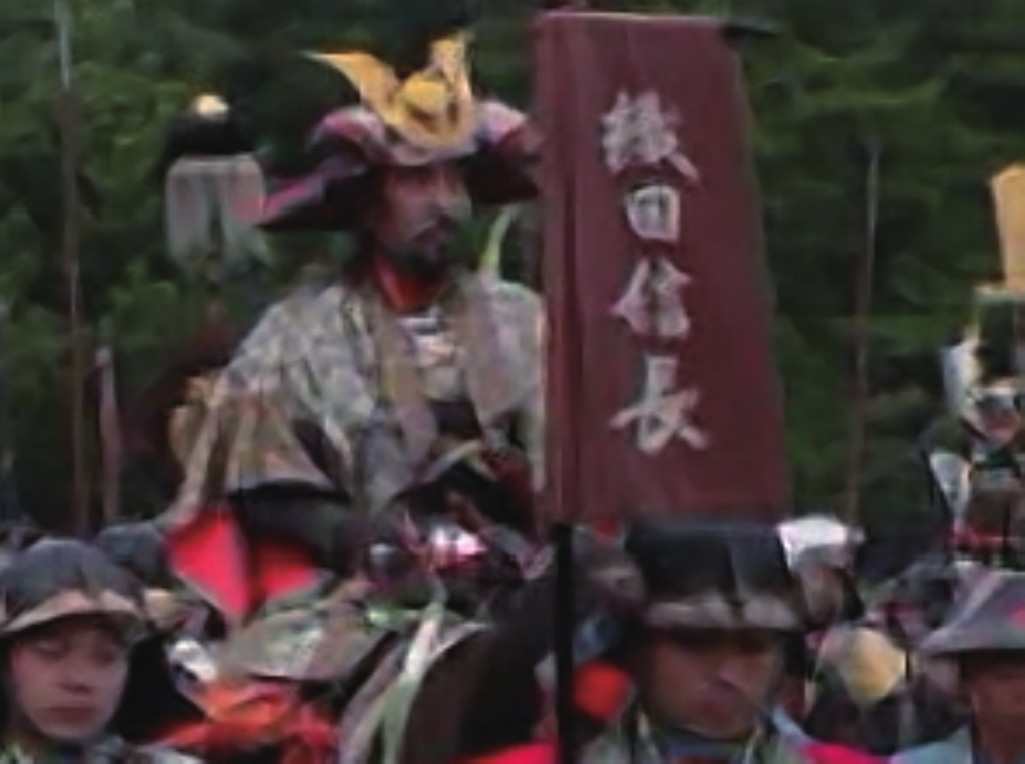}
        \end{center}
      \end{minipage}
	&
      % [6]proposed2
      \begin{minipage}{3.5cm}
        \begin{center}
          \includegraphics[clip, width=3.5cm]{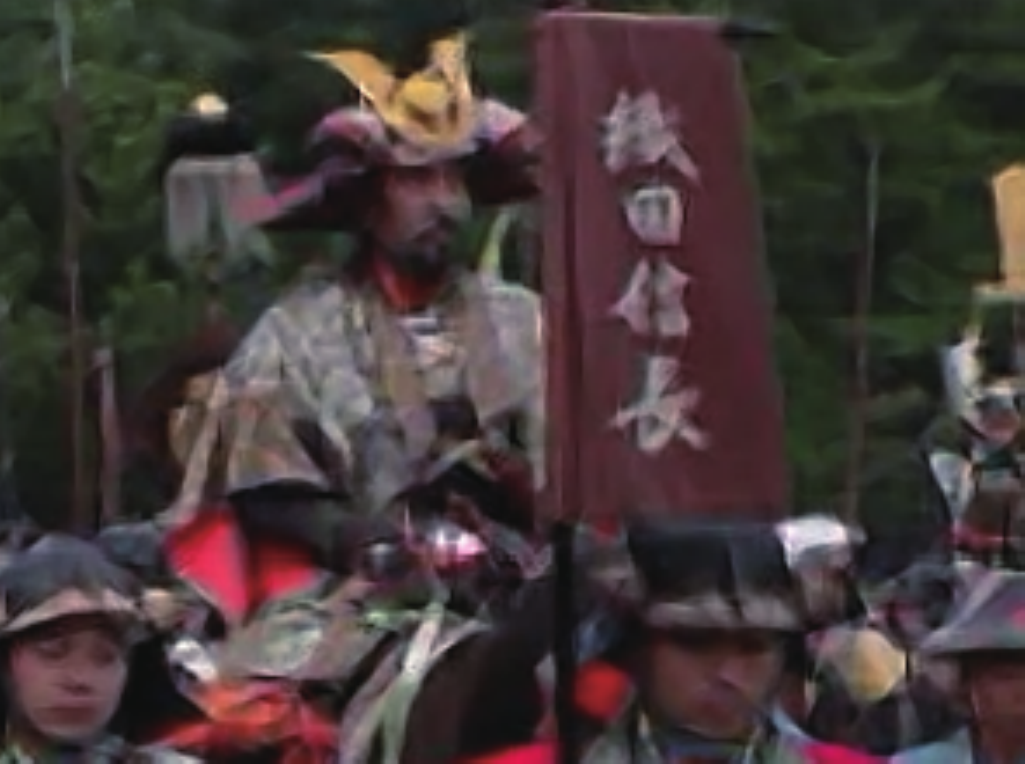}
        \end{center}
      \end{minipage}
	&
      % [7]Original
      \begin{minipage}{3.5cm}
        \begin{center}
          \includegraphics[clip, width=3.5cm]{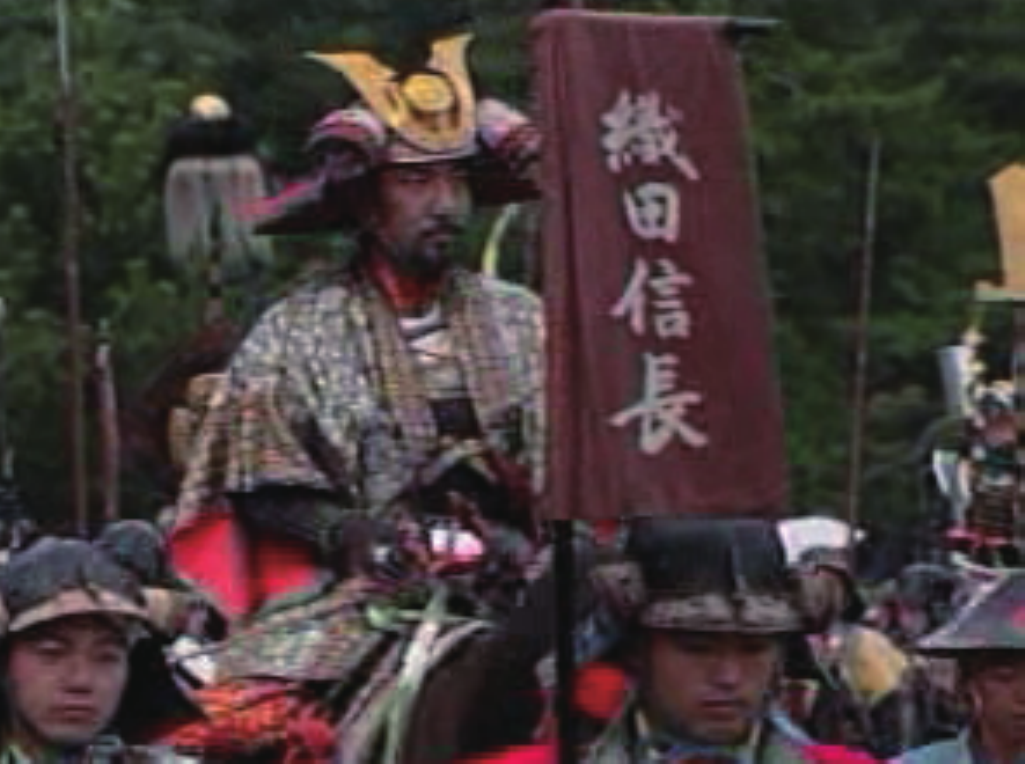}
        \end{center}
      \end{minipage}
		\\
	{\small MF-JDL}
	&
	{\small proposed($\delta=0.003$)}
	&
	{\small proposed($\delta=0.001$)}
	&
	{\small original image}
    \end{tabular}
\caption{Images estimated from LR observations (Samurai)}
    \label{r_samurai}
  \end{center}
\end{figure*}

\begin{figure*}[htbp]
  \begin{center}
    \begin{tabular}{cccc}
      % [1]Input
      \begin{minipage}{3.5cm}
        \begin{center}
          \includegraphics[clip, width=3.5cm]{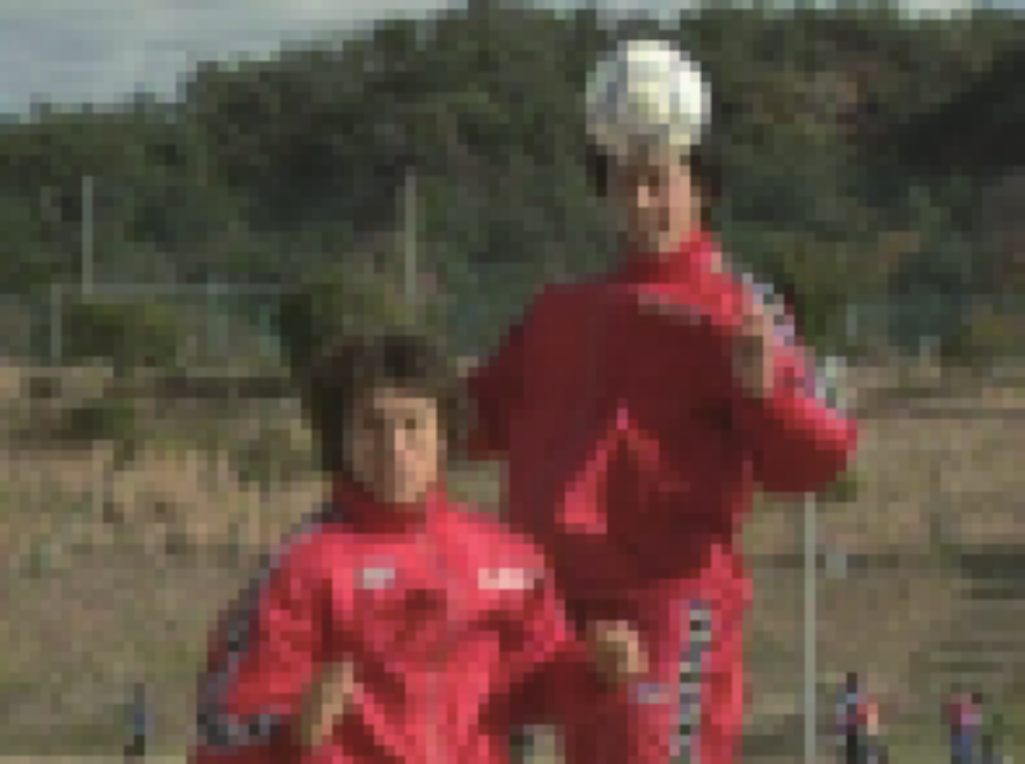}
        \end{center}
      \end{minipage}
	&
      %  [2]Bicubic
      \begin{minipage}{3.5cm}
        \begin{center}
          \includegraphics[clip, width=3.5cm]{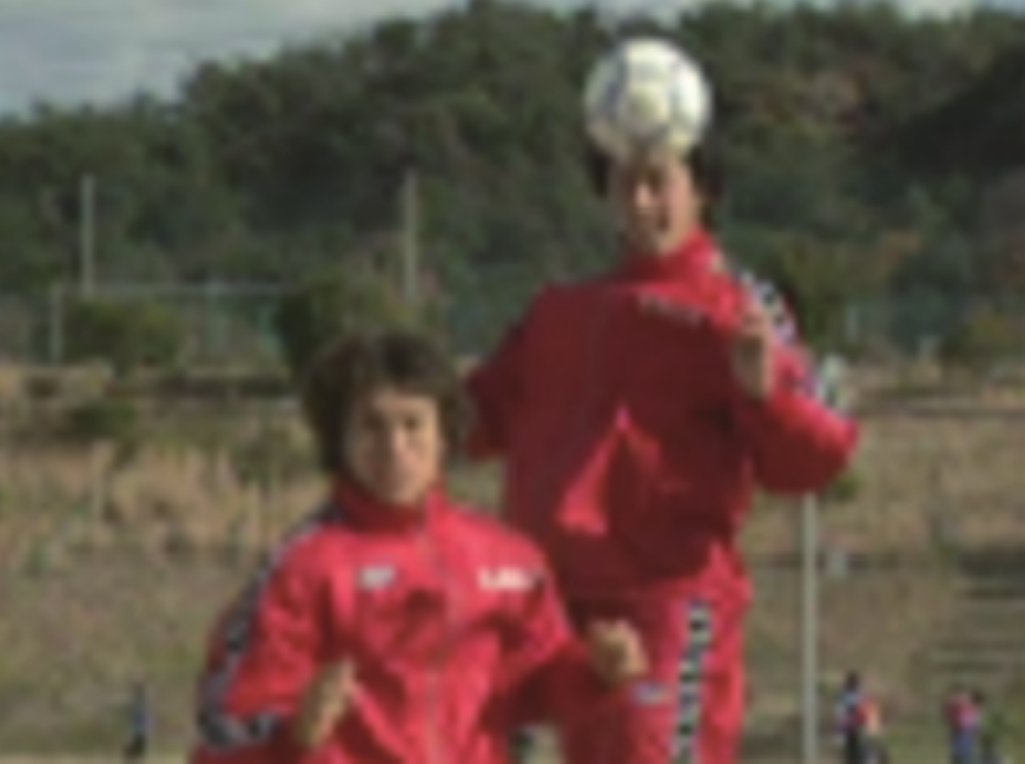}
        \end{center}
      \end{minipage}
	&
      % [3]Yang2010
      \begin{minipage}{3.5cm}
        \begin{center}
          \includegraphics[clip, width=3.5cm]{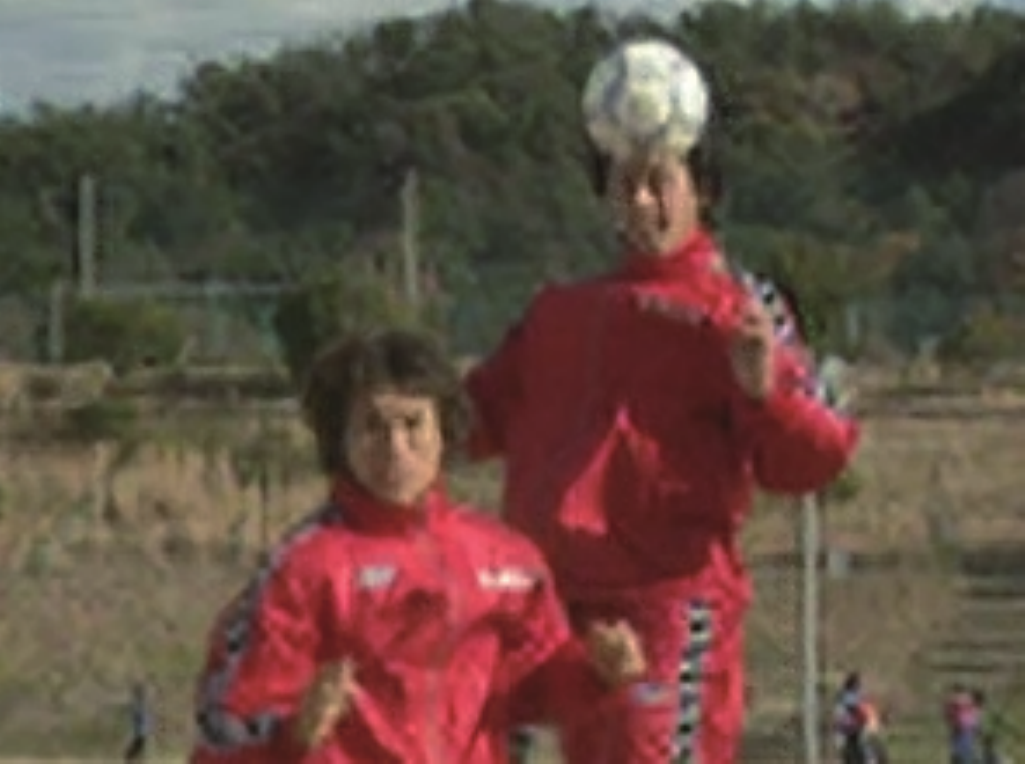}
        \end{center}
      \end{minipage}
	&
	   % [4]Zeyde2010
      \begin{minipage}{3.5cm}
        \begin{center}
          \includegraphics[clip, width=3.5cm]{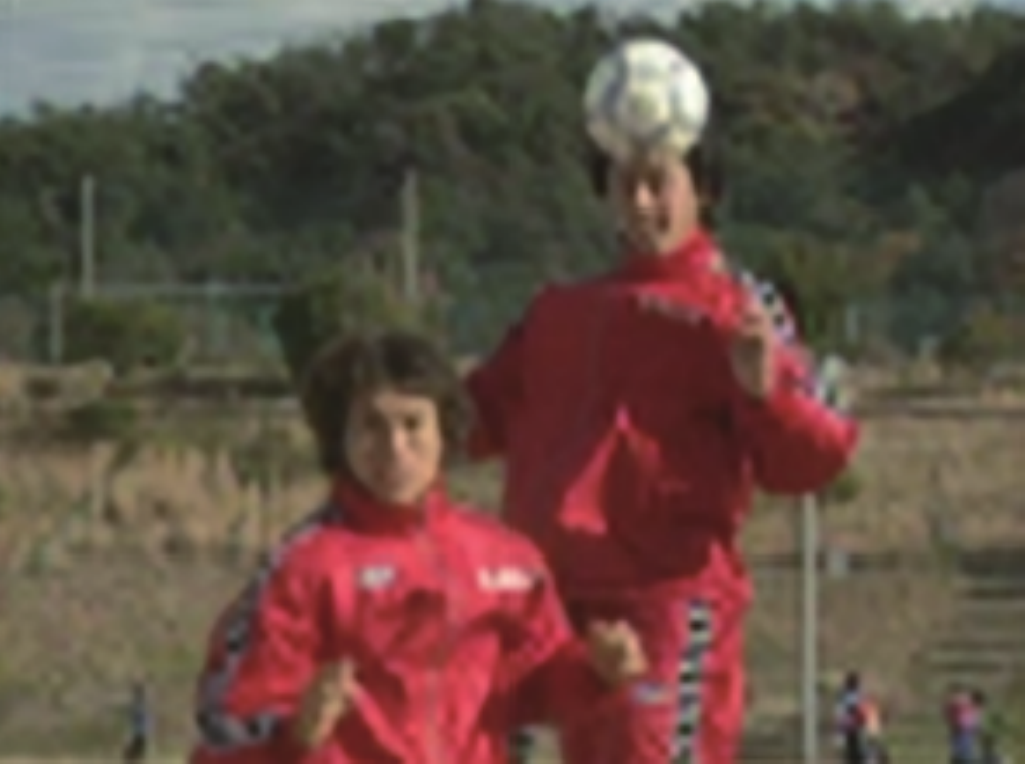}
        \end{center}
      \end{minipage}
	\\
	{\small input}
	&
	{\small Bi-cubic}
	&
	{\small SF-JDL}
	&
	{\small SF-L0}
		\\
	% [5]Wang2011
      \begin{minipage}{3.5cm}
        \begin{center}
          \includegraphics[clip, width=3.5cm]{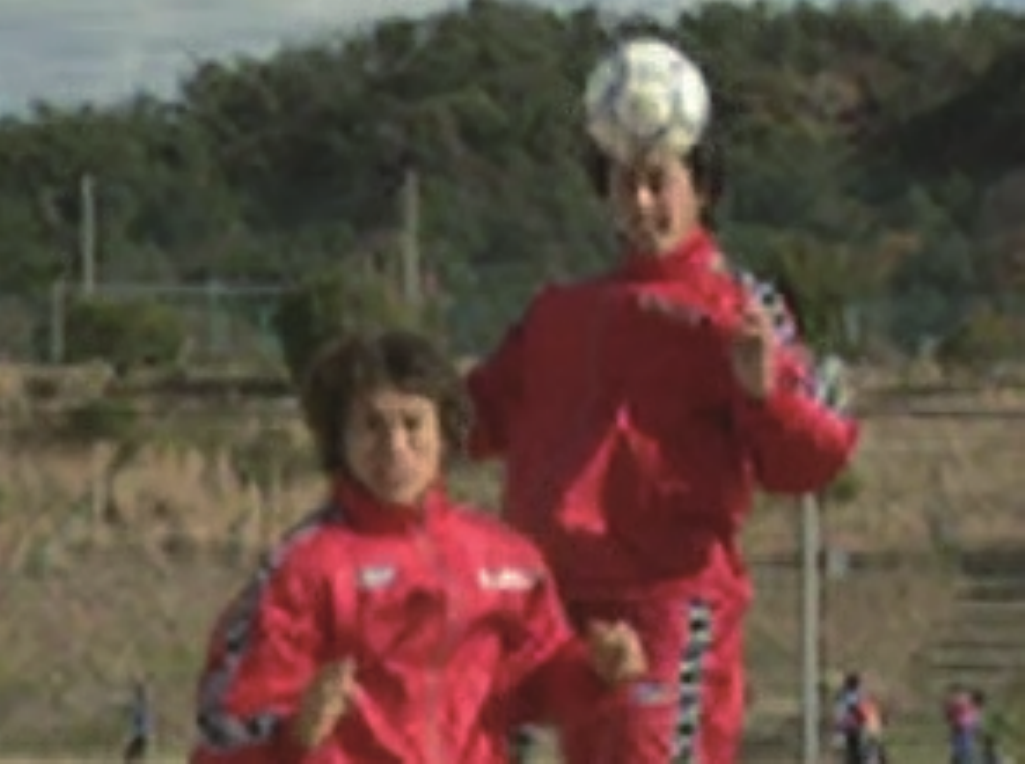}
        \end{center}
      \end{minipage}
	&
      % [6]proposed1
      \begin{minipage}{3.5cm}
        \begin{center}
          \includegraphics[clip, width=3.5cm]{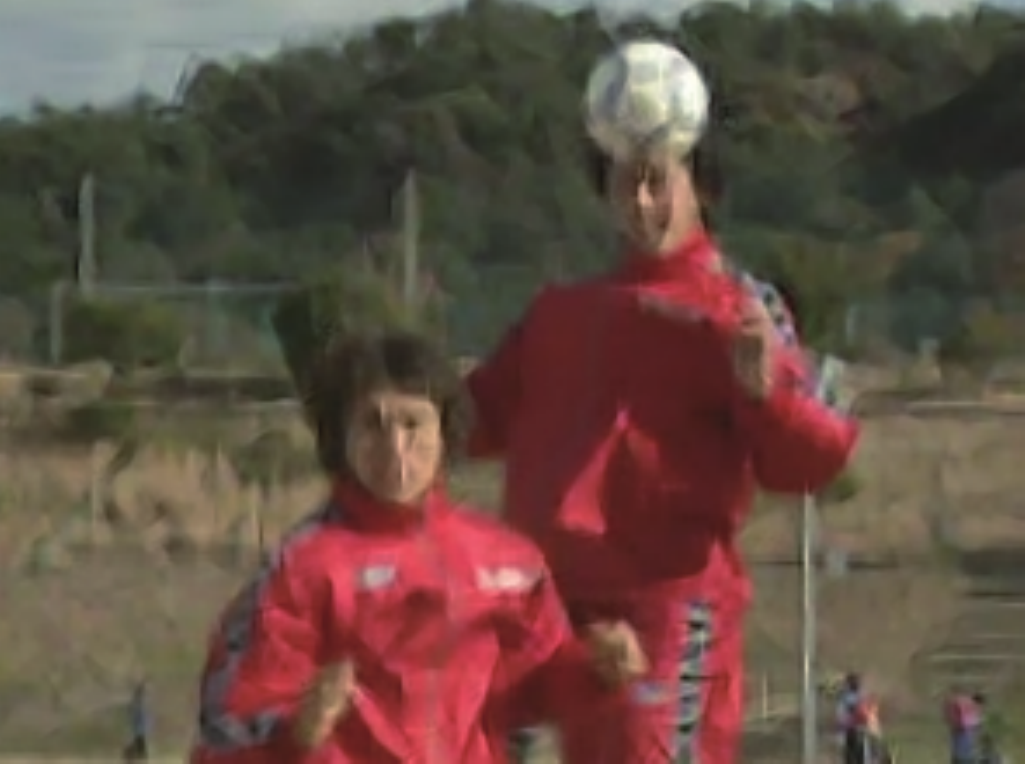}
        \end{center}
      \end{minipage}
	&
      % [7]proposed2
      \begin{minipage}{3.5cm}
        \begin{center}
          \includegraphics[clip, width=3.5cm]{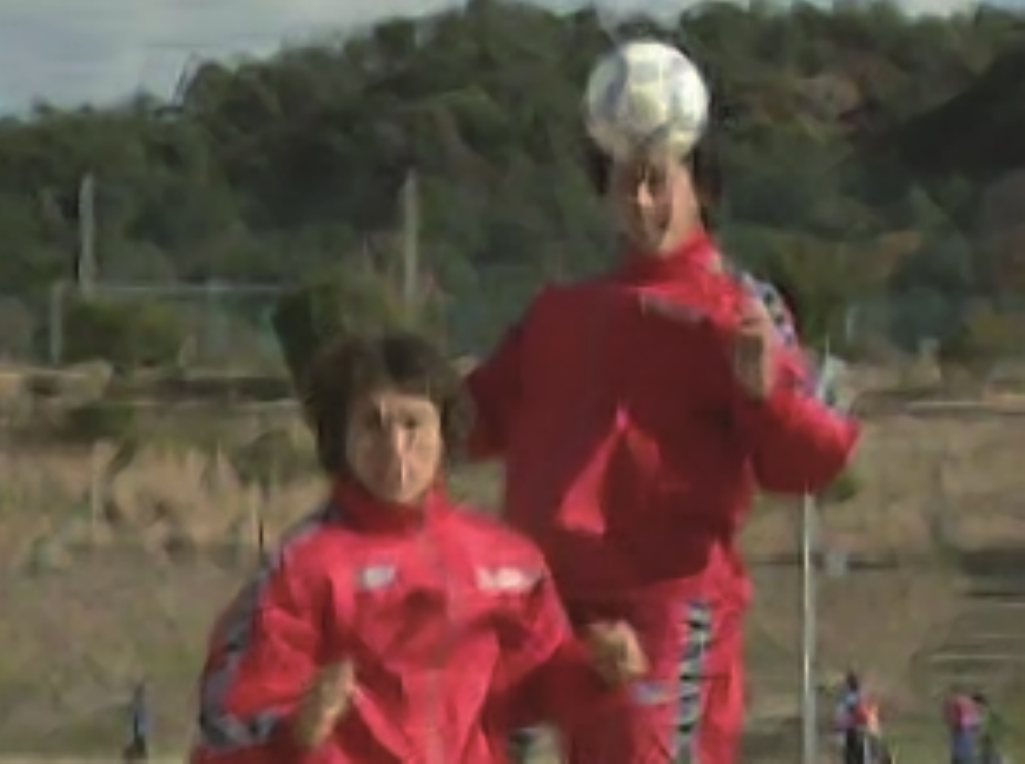}
        \end{center}
      \end{minipage}
	&
      % [8]Original
      \begin{minipage}{3.5cm}
        \begin{center}
          \includegraphics[clip, width=3.5cm]{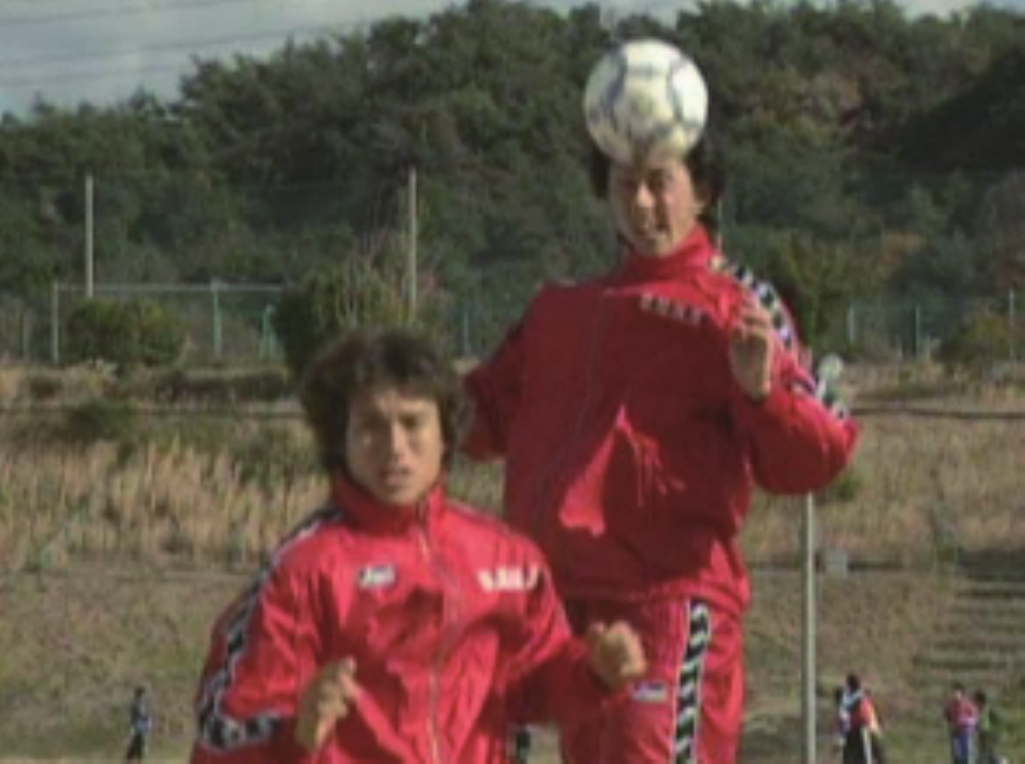}
        \end{center}
      \end{minipage}
		\\
	{\small MF-JDL}
	&
	{\small proposed($\delta=0.003$)}
	&
	{\small proposed($\delta=0.001$)}
	&
	{\small original image}
    \end{tabular}
\caption{Images estimated from LR observations (Heading)}
    \label{r_head}
  \end{center}
\end{figure*}

We use a gray-scale movie (MacArthur) and two different color movies
(Samurai and Heading) for evaluating the
performance of SR methods. We capture five consecutive images from each
movie, and make LR images according to the degradation model.
The obtained HR images using various SR methods are shown in 
Fig.~\ref{r_mac}, Fig.~\ref{r_samurai}, and Fig.~\ref{r_head}.

From Fig.~\ref{r_mac}, Fig.~\ref{r_samurai}, and
Fig.~\ref{r_head}, the HR images obtained by the proposed method
are clear and have distinct edges. By comparing to the original HR
images, we can see that the proposed method can reconstruct edges and
line-like features in the original images. On the other hand, 
 there are patches in which the proposed method can not clearly
 reconstruct textures like wrinkles and patterns in clothes.
Also, in the case of \lq\lq Heading\rq \rq, 
 the overall sharpness of the image is improved but undesirable
 artifacts are produced. This is possibly because the original movie
 contains abrupt shifts.

\section{Conclusion}
In this paper, we proposed a multi-frame SR method based
on sparse coding and sub-pixel accuracy block matching. The main contribution of the present paper is in
proposing a natural extension of the single-frame SR method based on the
sparse coding, with a novel combination of sub-pixel accuracy block
matching method and an LR atom generation from HR atoms. 
The proposed method takes the sub-pixel displacements into account for
the degradation process of HR images.
Another important contribution is that it can handle a
variable number of LR observations.
Among a set of observed LR images, the number of LR images actually
used for SR is automatically determined in each patch. 
There would be patches easy to estimate the displacement from the target patch, and 
difficult to estimate the displacement. When an LR image is composed of
those two different kinds of patches, this property of automatic patch
selection is useful because we can effectively utilize a subset of LR images
 where displacements are estimated with high confidences. 
When we deal with movies, it is difficult to estimate relative
displacements for the regions where
 objects move quickly, and it is easy to estimate the relative
displacements for the regions without such objects.
In general, for quality of a movie, the sharpness of objects at rest or
slowly moving are significant, and this property is advantageous to the
proposed method.

In our future work, we will investigate an SR method for movies based on sparse coding.
The proposed method is ready to be applied for movie SR by applying
frame by frame, though, the current method does not consider the
continuity between frames which should not be ignored for
movie SR.

In example-based sparse coding SR, the sparse coding procedure is
applied to certain kinds of features. Particularly, to ensure that the
computed coefficient fits the most relevant part of the LR image, most
of recent sparse coding based SR methods adopt high-pass filters on
both $\bf y$ and $\mathbf{D}_{l} \boldsymbol \alpha$. 
The proposed method does not resort to feature extraction because of
possible increase of computational costs.
 Finally, we are also pursuing a method to automatically determine the
threshold $\delta$ using the observed LR images.

\end{document}